\documentclass[10pt,twocolumn,letterpaper]{article}

\usepackage{cvpr}              

\usepackage{lipsum}
\usepackage{tcolorbox}
\usepackage{natbib}
\usepackage{stfloats}
\usepackage[accsupp]{axessibility}  

\usepackage[linesnumbered, ruled, vlined]{algorithm2e}
\newcommand{\algorithmstyle}[1]{\renewcommand{\algocf@style}{#1}}

\usepackage{algorithm2e}
\SetCommentSty{mycommfont}
\SetKwInput{KwInput}{Input}                
\SetKwInput{KwOutput}{Output}              
\SetKwInput{KwPrompt}{Prompts}             
\SetKw{KwStep}{step}
\SetKw{EndFor}{end for}
\SetKw{Require}{Require} 
\SetKw{Break}{Break}

\makeatletter
\newcommand{\removelatexerror}{\let\@latex@error\@gobble}
\makeatother

\usepackage{ifthen} 
\usepackage{etoolbox} 
\newboolean{showcomments}
\setboolean{showcomments}{true} 
\newrobustcmd{\sps}[1]{\ifthenelse{\boolean{showcomments}}{\textcolor{orange}{(\textbf{SP}: #1)}}{}}
\newrobustcmd{\mk}[1]{\ifthenelse{\boolean{showcomments}}{\textcolor{purple}{(\textbf{Minkyu}: #1)}}{}}
\newrobustcmd{\sas}[1]{\ifthenelse{\boolean{showcomments}}{\textcolor{purple}{(\textbf{Sahil}: #1)}}{}}

\usepackage{graphicx} 
\usepackage{multirow} 
\usepackage{pifont} 
\usepackage{longtable} 
\usepackage{comment} 

\usepackage{array}
\newcommand{\PreserveBackslash}[1]{\let\temp=\\#1\let\\=\temp}
\newcolumntype{C}[1]{>{\PreserveBackslash\centering}p{#1}}
\newcolumntype{R}[1]{>{\PreserveBackslash\raggedleft}p{#1}}
\newcolumntype{L}[1]{>{\PreserveBackslash\raggedright}p{#1}}
\usepackage{capt-of}
\DeclareCaptionType{prompt}[Prompt][List of Prompts]

\newcommand{\propset}{\mathcal{P}}
\newcommand{\prop}{p}
\newcommand{\spec}{\Phi}

\newcommand{\always}{$\Box$}
\newcommand{\eventually}{$\diamondsuit$}
\newcommand{\ournext}{$\mathsf{X}$}
\newcommand{\until}{$\mathsf{U}$}

\newcommand{\dtmc}{\mathcal{A}}
\newcommand{\va}{\mathcal{A}_{\mathcal{V}}}
\newcommand{\probsat}{\mathbb{P}[\va \models \Phi]}

\newcommand{\neusv}{\textit{NeuS-V}}

\newcommand{\ttv}{\mathcal{M}_{\text{T2V}}}
\newcommand{\vlm}{\mathcal{M}_{\text{VLM}}}
\newcommand{\ecdf}{f_{\text{ECDF}}}

\newcommand{\ourpivot}{\textit{PULS}}
\newcommand{\pivotf}{LM_{\text{PULS}}}
\newcommand{\ttp}{LM_{\text{T2P}}}
\newcommand{\tttl}{LM_{\text{T2TL}}}
\newcommand{\oprompt}{\theta^\star}
\newcommand{\dttp}{\mathfrak{D}_{\text{T2P}|\text{train}}}
\newcommand{\dttl}{\mathfrak{D}_{\text{T2TL}|\text{train}}}

\renewcommand{\propset}{\mathcal{P}}
\renewcommand{\prop}{p}

\renewcommand{\spec}{\Phi}

\renewcommand{\always}{$\Box$}
\renewcommand{\eventually}{$\diamondsuit$}
\renewcommand{\ournext}{$\mathsf{X}$}
\renewcommand{\until}{$\mathsf{U}$}

\renewcommand{\dtmc}{\mathcal{A}}
\renewcommand{\va}{\mathcal{A}_{\mathcal{V}}}

\renewcommand{\neusv}{\textit{NeuS-V}}

\renewcommand{\ttv}{\mathcal{M}_{\text{T2V}}}
\renewcommand{\vlm}{\mathcal{M}_{\text{VLM}}}

\renewcommand{\ecdf}{f_{\text{ECDF}}}

\renewcommand{\ourpivot}{\textit{PULS}}
\renewcommand{\pivotf}{LM_{\text{PULS}}}
\renewcommand{\ttp}{LM_{\text{T2P}}}
\renewcommand{\tttl}{LM_{\text{T2TL}}}
\renewcommand{\oprompt}{\theta^\star}
\renewcommand{\dttp}{\mathfrak{D}_{\text{T2P}|\text{train}}}
\renewcommand{\dttl}{\mathfrak{D}_{\text{T2TL}|\text{train}}}

\definecolor{cvprblue}{rgb}{0.21,0.49,0.74}
\usepackage[pagebackref,breaklinks,colorlinks,allcolors=cvprblue]{hyperref}

\def\paperID{18233} 
\def\confName{CVPR}
\def\confYear{2025}

\title{Neuro-Symbolic Evaluation of Text-to-Video Models using Formal Verification}

\author{
  \textbf{S P Sharan}\textsuperscript{*} \qquad 
  \textbf{Minkyu Choi}\textsuperscript{* \dag} \qquad 
  \textbf{Sahil Shah} \\
  \textbf{Harsh Goel} \qquad 
  \textbf{Mohammad Omama} \qquad 
  \textbf{Sandeep Chinchali} \\
  \\
  The University of Texas at Austin, United States \\
}

\begin{document}


\twocolumn[{
\maketitle
\begin{center}
    \captionsetup{type=figure}

    \includegraphics[width=\linewidth]{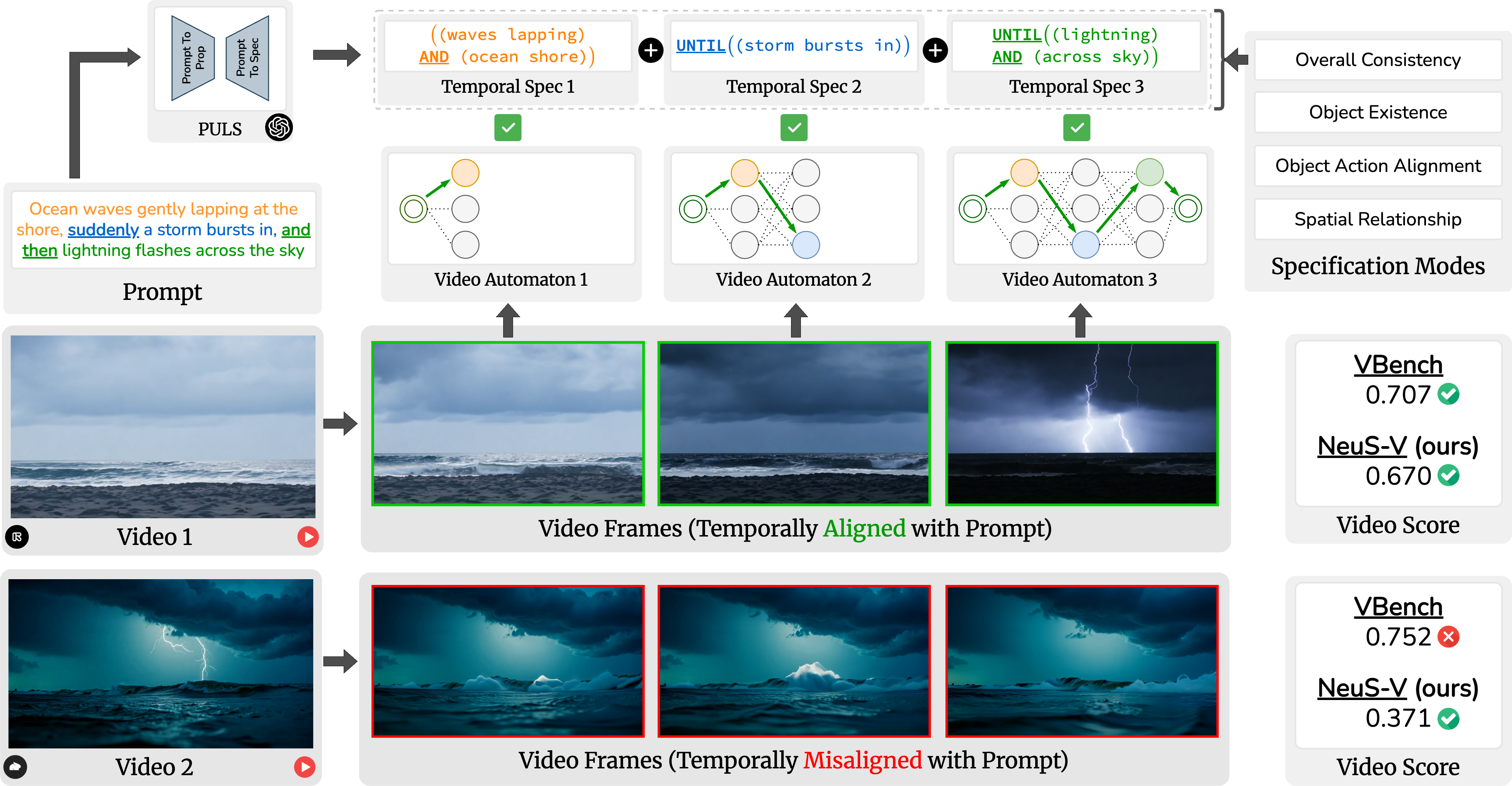}

    

    \captionof{figure}{\textbf{Current generative video evaluation methods struggle with temporal fidelity.} \neusv converts prompts into Temporal Logic specifications and formally verifies them against a video automaton. The upper video aligns with the prompt's temporal sequencing, while the lower video, despite being visually appealing, fails to do so. Unlike VBench, \neusv{} effectively differentiates between them.}
    \label{fig:teaser}
\end{center}}]
\footnotetext{
    \hangindent=1em
    *Equal contribution (Order determined by coin toss). \\
    \hangindent=1em
    \dag Corresponding author: {\tt\small minkyu.choi@utexas.edu}
}




\begin{abstract}
    Recent advancements in text-to-video models such as Sora, Gen-3, MovieGen, and CogVideoX are pushing the boundaries of synthetic video generation, with adoption seen in fields like robotics, autonomous driving, and entertainment. As these models become prevalent, various metrics and benchmarks have emerged to evaluate the quality of the generated videos. However, these metrics emphasize visual quality and smoothness, neglecting temporal fidelity and text-to-video alignment, which are crucial for safety-critical applications. To address this gap, we introduce \(\neusv\), a novel synthetic video evaluation metric that rigorously assesses text-to-video alignment using neuro-symbolic formal verification techniques. Our approach first converts the prompt into a formally defined Temporal Logic (TL) specification and translates the generated video into an automaton representation. Then, it evaluates the text-to-video alignment by formally checking the video automaton against the TL specification. Furthermore, we present a dataset of temporally extended prompts to evaluate state-of-the-art video generation models against our benchmark. We find that \(\neusv\) demonstrates a higher correlation by over $5 \times$ with human evaluations when compared to existing metrics. Our evaluation further reveals that current video generation models perform poorly on these temporally complex prompts, highlighting the need for future work in improving text-to-video generation capabilities. 
    We open-source our benchmark, code, and dataset at \href{https://utaustin-swarmlab.github.io/NeuS-V/}{\textnormal{\texttt{utaustin-swarmlab.github.io/NeuS-V}}}.
\end{abstract}

\section{Introduction}

    Imagine that we need to stress test or retrain a motion planning system for autonomous driving (AD) by simulating a scenario where ``a truck appears in frame $10$ and veers in front of me onto my lane within 2 seconds''. With recent developments in text-to-video (T2V) generative models, generating synthetic videos for such scenarios offers great potential for improving such engineering systems. However, for these synthetic videos to be effective, they must align with the desired prompt across three key dimensions: \ding{202}~\textit{semantics}, \ding{203} \textit{spatial relationships}, and \ding{204} \textit{temporal coherence}. For example, if the synthetic video either inaccurately generates the truck’s movement in front of the autonomous vehicle or fails to follow the correct lane-change timing, it could mislead the motion planning system during retraining, resulting in incorrect decisions in critical situations.

    Various evaluation metrics have emerged to identify well-aligned synthetic videos to the prompt from recently developed T2V models \citep{ho2020denoising, esser2023structure, blattmann2023align, zhang2023i2vgen, li2024reward}. While many video evaluation tools \citep{liu2024fetv, liu2024evalcrafter, huang2024vbench} focus on visual quality, some approaches use vision-language models \citep{sun2024t2v, wu2024towards} or video captioning models \citep{chivileva2023measuring} to evaluate the semantic alignment between the generated video and the original prompt. More notably, VBench \cite{huang2024vbench} evaluates videos across multiple dimensions and categories and is recently being used in leaderboards\footnote{\href{https://huggingface.co/spaces/Vchitect/VBench_Leaderboard}{https://huggingface.co/spaces/Vchitect/VBench\_Leaderboard}} to evaluate T2V models. However, naive evaluation of a video generated by one neural network using another is neither \textit{rigorous} nor \textit{interpretable}. Moreover, current evaluation metrics still miss critical spatio-temporal aspects that are intrinsic to video data (See \cref{fig:teaser}). 
    
    Our key insight is to leverage the recent advancements in video understanding systems that utilize structured symbolic verification methods with neural networks \cite{choi2025towards}. We introduce \textbf{Neuro-Symbolic Verification (\neusv{})}, a novel T2V model evaluation method that integrates neural network outputs with structured \textit{symbolic verification} through \textit{temporal logic (TL) specifications}.  First, we leverage Vision-Language Models (VLMs) to interpret a video's spatial and semantic content. Second, we capture temporal relationships across these identified semantics by processing frames sequentially. Finally, we construct an automaton representation of the synthetic video which establishes a structured temporal sequence of the video events. This structure ensures that the specified TL requirements are met, resulting in a \textit{rigorous} and \textit{interpretable} video evaluation.
        
    Furthermore, we introduce a dataset of temporally extended prompts for benchmarking T2V models' abilities to faithfully exhibit temporal fidelity and accurately reflect event sequences. We plan to open-source \neusv{}, our prompt suite, and human evaluation results along with a public leaderboard of state-of-the-art (SOTA) T2V models.
    
\section{Related Work}  

    \paragraph{Evaluation of Text-to-Video Models.} 
        The recent surge of T2V models has created a demand for specialized metrics to evaluate the quality and coherence of generated videos. Some preliminary works in this field utilize metrics such as FID, FVD, and CLIPSIM \cite{shin2024lostmelodyempiricalobservations, liu2023fetvbenchmarkfinegrainedevaluation, jain2024peekaboointeractivevideogeneration, bugliarello2023storybenchmultifacetedbenchmarkcontinuous, hu2022makemovecontrollableimagetovideo, yu2023celebvtextlargescalefacialtextvideo}, but these approaches struggle with prompts that require complex reasoning and logic. Recent works \citep{kou2024subjectivealigneddatasetmetrictexttovideo, liu2024evalcrafterbenchmarkingevaluatinglarge, zhang2024benchmarkingaigcvideoquality, Li_2024_CVPR} rely on Large Language Models (LLMs) to break down prompts into Visual Question Answering (VQA) pairs, which are then scored by a VLM. Few approaches also utilize vision transformers and temporal networks \cite{huang2023t2i, ji2024t2vbench, feng2024tc, chu2024sora}. Some works, including EvalCrafter \cite{liu2024evalcrafter}, FETV \cite{liu2024fetv}, T2V-Bench \cite{feng2024tc}, and others \cite{he2024videoscorebuildingautomaticmetrics, Ji_2024_CVPR}, incorporate an ensemble of various visual quality metrics. Of these, VBench \citep{huang2024vbench} is rising to be the de-facto benchmark in the field, evaluating across numerous dimensions and categories. However, all current evaluation methods emphasize \textit{visual quality}, disregarding \textit{temporal ordering} of videos. Although some \textit{claim} to perform temporal evaluation, they often focus on aspects such as playback (e.g. slow motion, timelapse) or movement speed rather than true temporal reasoning and also lack rigor. In contrast, \neusv{} addresses these shortcomings by formally verifying temporal fidelity through TL specification over the automaton representation of synthetic videos. 

    \paragraph{Video Event Understanding.} 
        Recent methods typically rely on perception models \citep{fan2021pytorchvideo, pradhyumna2021graph} or VLMs \citep{zhang2023video, maaz2023video} to analyze events within a video. Although these models can detect objects and actions effectively, they do not guarantee a reliable understanding of the video's temporal dynamics. \citet{yang2023specification} focuses on representing a video as a formal model, specifically a Discrete Markov Chain, providing a more structured approach to capture events across time. An application of this is NSVS-TL \citep{choi2025towards}, which uses a neuro-symbolic approach to find the scenes of interest. Along similar lines, $\neusv$ leverages VLMs and formal verification to enhance comprehension of temporal relationships in generated videos and assess video quality, thus establishing a reliable foundation for video evaluation.

\section{Preliminaries}
\label{sec:preliminaries}
    In the following sections, we present a running example to help illustrate our approach. Let's say we generate an autonomous driving video using T2V models with the prompt -- ``A car drives down a road on a clear sunny day, interacting with cyclists who signal turns to avoid obstacles".
    
    \paragraph{Temporal Logic.}
        In short, TL is an expressive formal language with logical and temporal operators \citep{Temporal-and-Modal-Logic, Manna}. A TL formula consists of three parts: \ding{202} a set of atomic propositions, \ding{203} first-order logic operators, and \ding{204} temporal operators. Atomic propositions are indivisible statements that can be \texttt{True} or \texttt{False}, and are composed to construct more complex expressions. The first-order logic operators include AND (\(\wedge\)), OR (\(\vee\)), NOT (\(\neg\)), IMPLY (\(\Rightarrow\)), etc., and the temporal operators consist of ALWAYS (\always), EVENTUALLY (\eventually), NEXT (\ournext), UNTIL (\until), etc. 

        The atomic propositions $\propset$, and TL specifications $\Phi$ of our example are:
        \begin{equation}
            \label{eq:running_example_tl_spec}
            \begin{array}{rl}
                \propset &= \{\text{car driving}, \ \text{clear day}, \ \text{cyclist signals turn}, \\ 
                   &\quad \ \ \text{cyclist turns}, \ \text{cyclist avoids obstacle}\}, \\ 
                \Phi &= \Box \Big((\text{car driving} \wedge \text{clear day}) \wedge (\text{cyclist signals turn})
            \end{array}
        \end{equation}
    
        This TL specification illustrates that if a car drives on a clear day \textit{and} a cyclist signals to turn, it \textit{implies} that \textit{eventually} a cyclist will turn and avoid an obstacle. 

\begin{figure}[ht]
            \centering
            \includegraphics[width=\linewidth]{
            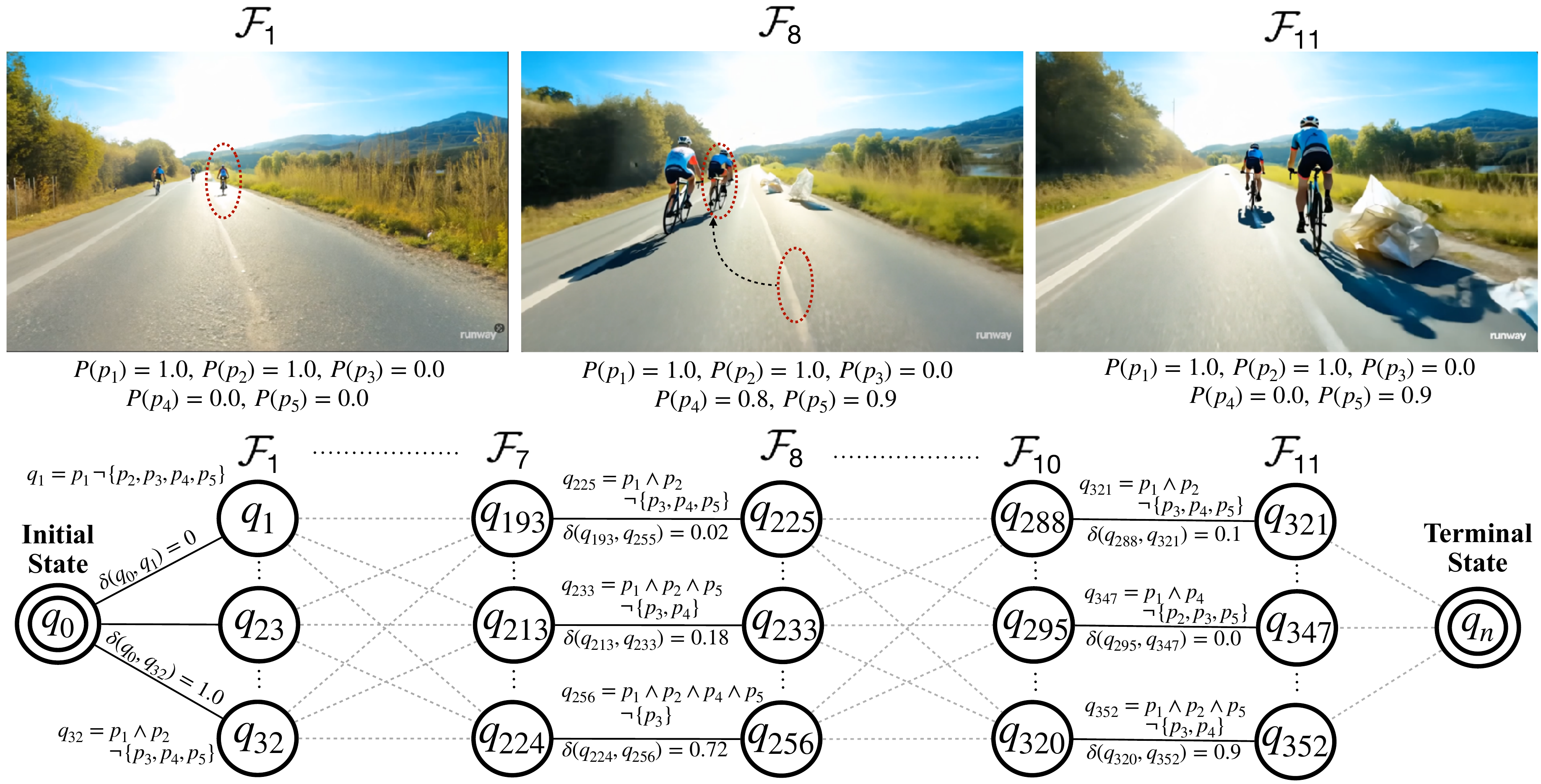}
            \caption{\textbf{Video automaton from the running example}. Above is an automaton of a video generated by the Gen-3 model constructed with a TL specification (See \cref{eq:running_example_tl_spec}). Every state $q_t$ from a frame $\mathcal{F}$ is labeled and has incoming and outgoing transition probabilities $\delta(q,q') \in [0,1]$. For example, in frame $\mathcal{F}_8$, we have probabilities $P(p_4)=0.8$ and $P(p_5)=0.9$, where $p_4$ represents the atomic proposition ``cyclist turns'', and $p_5$ represents ``cyclist avoids obstacle''. These probabilities are assigned because the cyclist (red-dotted circle) has turned left to avoid obstacles on the road. In the state $q_{256}$, we have an incoming probability $0.72 = P(p_1) \times P(p_2) \times P(p_4) \times P(p_5) \times (1-P(p_3)) $ from the previous state $q_{224}$, where that label is true for $p_1,p_2,p_4$, and $p_5$, and false for $p_3$ denoted as $\neg\{p_3\}$.} 
            \label{fig:running_example_dtmc}
        \end{figure}

    \paragraph{Discrete-Time Markov Chain.}
        Abbreviated DTMC, it is used to model stochastic processes where transitions between states occur at discrete time steps \citep{norris1998markov, kemeny1960finite}. We use a DTMC to model synthetic videos since the sequence of frames is \textit{discrete} and \textit{finite}, as shown in \cref{fig:running_example_dtmc}. The DTMC is defined as a tuple $\dtmc=(Q,q_0,\delta,\lambda)$, where $Q$ is a finite set of states, $q_0 \in Q$ is the initial state, $\lambda:Q \to 2^{|\propset]}$ is the label function, $\propset$ is the set of atomic propositions and $\delta:Q \times Q \to [0,1]$ is the transition function. For any two states $q,q' \in Q$, the transition function $\delta(q,q') \in [0,1]$ gives the probability of transitioning from $q$ to $q'$. For each state, the probabilities of all outgoing transitions sum to 1. 

    \paragraph{Formal Verification.}
        This provides formal guarantees that a system meets a desired specification \citep{clarke1999model, huth2004logic}. It requires a formal representation of the system, such as a finite-state automaton (FSA) or a Markov Decision Process (MDP). From the DTMC, we define a \textit{path} as a sequence of states from the initial state, e.g., $q_0q_1(q_2)^{\omega}$, where $\omega$ indicates infinite repetition. A \textit{trace} is the sequence of state labels denoted as $\lambda(q_0)\lambda(q_1)\lambda(q_2) \cdots \in (2^{|\propset|})^{\omega}$ that captures events over time. Next, we apply probabilistic model checking \citep{Baier2008} to compute the probability $\mathbb{P}[\mathcal{A} \models \Phi]$ which gives the satisfaction probability that the \textit{trace} starting from the initial state satisfies the TL specification $\Phi$. Through such formal representations, we evaluate synthetic videos.

\section{Methodology}

    \begin{figure}[ht]
        \centering
        \includegraphics[width=\linewidth]{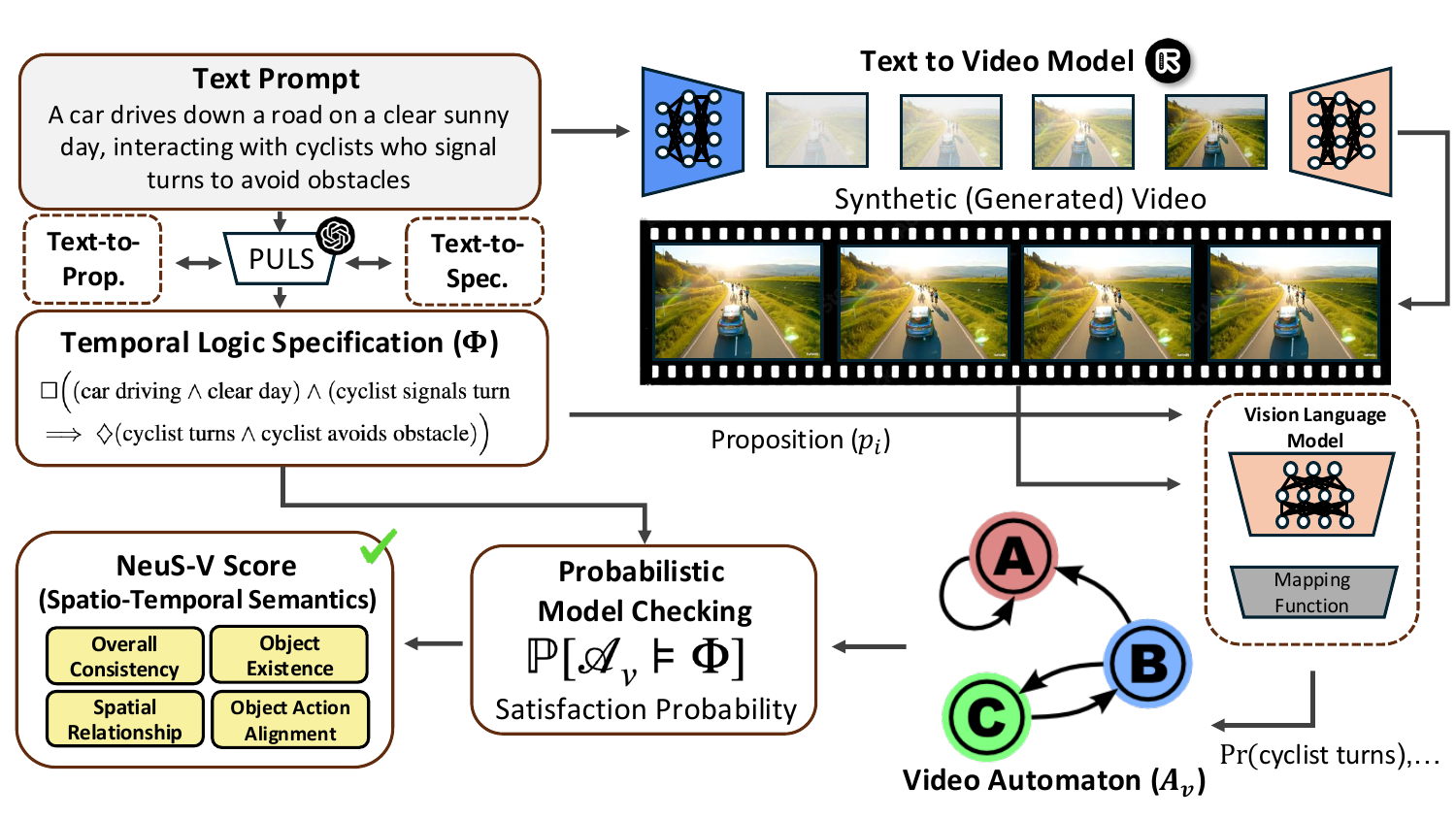}
        \caption{\textbf{Spatio-temporal and semantic measurements between a text prompt and a video by NeuS-V}. We first decompose the text prompt to TL specification $\spec$, then transform the synthetic video into an automaton representation $\va$. Finally, we calculate the satisfaction probability by probabilistically checking the extent to which $\va$ satisfies $\spec$.} 
        \label{fig:flowchart}
    \end{figure}

    Given a synthetic video $\mathcal{V}$ from T2V model $\mathcal{M_\text{T2V}}: T \to \mathcal{V}$, where $T$ is a text prompt, we aim to compute an evaluation score $S: T \times \mathcal{V} \to \mathbb{R}^m$ for each evaluation mode $m \in M$ = \{object existence, spatial relationship, object action alignment, overall consistency\}. We introduce \(\neusv\), a novel method for evaluating T2V models using a neuro-symbolic formal verification framework. \(\neusv\) evaluates synthetic video across four different evaluation modes: \textbf{object existence}, \textbf{spatial relationship}, \textbf{object and action alignment}, and \textbf{overall consistency} following these steps:
    
    \begin{itemize}
        \item \textbf{Step 1}: We translate the text prompt into a set of atomic propositions and a TL specification through Prompt Understanding via temporal Logic Specification ($\ourpivot$).
        \item \textbf{Step 2}: We obtain a semantic confidence score for each atomic proposition using a VLM applied to each sequence of frames.
        \item \textbf{Step 3}: We construct a video automaton representation for the synthetic video.
        \item \textbf{Step 4}: We compute the satisfaction probability by formally verifying the constructed video automaton against the TL specification.
        \item \textbf{Step 5}:  Finally, we calibrate the satisfaction probability by mapping it to the final evaluation score from $\neusv$.
    \end{itemize}

    \subsection{Prompt Understanding via Temporal Logic Specification}
        We propose \textbf{Prompt Understanding via temporal Logic ($\ourpivot$)}, a novel method that facilitates an efficient and accurate conversion of text prompts to TL specifications. It preserves all semantic details of the prompt to formulate its TL specification. In contrast to previous works that focus solely on formulating relationships between singular objects, isolated events \citep{Fuggitti_Chakraborti_2024, brunello_et_al:LIPIcs.TIME.2019.17, chen2024nl2tltransformingnaturallanguages}, or encoding spatial information \citep{liu2023lang2ltl, pan2023dataefficientlearningnaturallanguage}, our method, $\ourpivot$, fully captures both temporal and spatial details. By incorporating four evaluation modes, we provide a more comprehensive assessment across spatio-temporal dimensions.
    
        \paragraph{Evaluation Mode.}
            $\ourpivot$ compiles the text prompt into the different set of proposition $\propset$ and TL specifications $\spec$ across four evaluation modes:
            
            \begin{itemize}
                \item \textbf{Object Existence:} This mode evaluates the existence of objects specified in the text prompt. For instance, in our running example (\cref{sec:preliminaries}), the TL specification for the object existence mode can be defined as $\spec=\diamondsuit(\text{car} \ \wedge \ \text{cyclist} \ \wedge \ \text{obstacle}).$ 
            
                \item \textbf{Spatial Relationship:} This evaluation mode captures the spatial relationships among existing objects. In our running example, since we are interested in a cyclist signaling and then making a turn, it implies that the cyclist is in front of the car. Hence, the TL specification is defined as $\spec = \Box(\text{cyclists are in front of a car}) \ \wedge \ \diamondsuit(\text{obstacles are next to cyclists})$. 
            
                \item \textbf{Object and Action Alignment:}  It ensures that the objects in the generated video perform the actions specified in the text prompt. For example, the TL specification can be defined as $\spec = \diamondsuit(\text{cyclist signals turn} \wedge \text{cyclist turns} \ $\until$ \ \text{cyclist avoids obstacle})$.
            
                \item \textbf{Overall Consistency:} Finally, we assess the overall semantic and temporal coherence of events in the generated video, ensuring alignment with the full-text prompt (see the example in \cref{eq:running_example_tl_spec}).
            \end{itemize} 
    
        \paragraph{Functional Modules.}
            $\ourpivot$ is a two-step algorithm powered by LLMs. Given the text prompt $\mathcal{T}$ that is used for video generation, it is translated into $\propset$ and $\spec$ for any given evaluation mode $m \in M$.  This translation process is defined as $LM_{\text{$\ourpivot$}} : \mathcal{T} \times M \to (\propset, \spec).$
        
            $\ourpivot$ consists of two modules, Text-to-Proposition (T2P) and Text-to-TL (T2TL). These use the optimized prompts $\oprompt_{T2P}$ and $\oprompt_{T2TL}$ respectively. Each optimized prompt includes carefully selected few-shot examples from a compiled training dataset $\dttl$, that maximize the accuracy of having correct $\propset$ and $\spec$ given $m \in M$. 
            
            \begin{enumerate}
                \item \textbf{Text-to-Proposition Module}: This module extracts $\propset$ from $\mathcal{T}$ given $M$, using $\oprompt_{\text{T2P}}$, and is defined as: 
                
                \begin{equation} 
                    \label{eq:ttp}
                    \ttp : \mathcal{T} \times M \xrightarrow{\oprompt_{\text{T2P}}} \propset.
                \end{equation}
                
                The optimized prompt $\oprompt_{\text{T2P}}$ is obtained from $\dttl$, a set of $\mathcal{N}$ selected examples of text $\mathcal{T}_i$ and propositions $\propset_i$ pairings :
                
                \begin{equation} 
                    \label{eq:dttp}
                    \dttp = \{ (\mathcal{T}_i, \propset_i) \}_{i=1}^{\mathcal{N}}, \quad \dttp \subset \mathfrak{D}_{\text{train}}.
                \end{equation}
                
                \item \textbf{Text-to-TL Specification Module}: This module generates $\spec$ from $\mathcal{T}$, $\propset$, and $M$ using the optimized prompt $\oprompt_{\text{T2TL}}$, and is defined as: 
                
                \begin{equation} 
                    \label{eq:tttl}
                    \tttl : \mathcal{T} \times \propset \times M \xrightarrow{\theta^\star_{\text{T2TL}}} \spec. 
                \end{equation}
                
                The optimized prompt $\oprompt_{\text{T2TL}}$ is obtained from $\dttl$ with $\mathcal{N}$ examples pairs of text  $\mathcal{T}_i$, propositions $\propset_i$, and corresponding TL specifications $\spec_i$:
                
                \begin{equation} 
                    \label{eq:dttl}
                    \dttl = \{ (\mathcal{T}_i, \propset_i,\spec_i)  \}_{i=1}^{\mathcal{N}}, \quad \dttl \subset \mathfrak{D}_{\text{train}}.
                \end{equation}
            
            \end{enumerate}
            We use MIPROv2 \citep{opsahlong2024optimizinginstructionsdemonstrationsmultistage} with DSPy \citep{khattab2023dspycompilingdeclarativelanguage} to ensure that the $\mathcal{N}$ few-shot examples in the dataset $\mathfrak{D}_{\text{train}}$ are sufficient in maximizing the accuracy of each module. We detail the optimization process in the Appendix.

    \subsection{Semantic Score from Neural Perception Model}    
    \label{subsec:semantic_score_from_neural_perception_model}
        Given $\propset,\spec=\pivotf(T)$, where $T$ is the text prompt used for the synthetic video $\mathcal{V}$, we first obtain the semantic confidence score $C \in [0, 1]$ for each atomic proposition $\prop_i \in \propset$ using a VLM $\vlm: \prop \times \mathcal{F} \to C$, where $\mathcal{F}$ represents a sequence of frames. 
    
        We use the VLM to interpret semantics \citep{metallama32v, internvl2024v2, liu2023videollava, zhang2023video} and extract confidence scores from $\mathcal{F}$ based on the text query $T$ . We pass each $\prop_i \in \propset$ along with the prompt to the VLM and calculate the token probability for the output response, which is either \texttt{True} or \texttt{False}. To calculate the token probability, we retrieve logits for the response tokens and compute the probability of that token after applying a softmax. Next, the semantic confidence score is the product of these probabilities as follows:
        \begin{equation}
            \label{eq:vlm}
            \begin{split}
                c_i &= \vlm(p_i, \mathcal{F}) \\
                    &= \prod_{j=1}^{k} P\left( t_j \mid \prop_i, \mathcal{F}, t_1, \dots, t_{j-1} \right) \  \forall \, p_i \in \propset,
            \end{split}
        \end{equation}
        where $(t_1,\dots,t_k)$ is the sequence of tokens in the response. The probability of the token at $j$'th response token is $P(t_j | \cdot) = \frac{e^{l_{j,t_j}}}{\sum_{z}e^{l_{j,z}}}$, where $l_{j,*}$ is the logit vector, and $l_{j,{t_j}}$ is the logit corresponding to token $t_j$. Finally, we calibrate the VLM to map the confidence to the probability using the semantic score mapping function $c^\star_i \leftarrow f_{\text{VLM}}(c_i;\gamma_{fp})$, where $c^\star_i$ is the calibrated confidence and $\gamma_{fp}$ is the false positive threshold. We provide additional details about the calibration process and the prompt design for the semantic detector VLM in the Appendix.

\vspace{2ex}
{
\begingroup
\removelatexerror
\resizebox{0.85\linewidth}{!}{%

    \begin{algorithm}[H]
        \footnotesize
        \DontPrintSemicolon
        \SetKwInOut{KwInput}{Input}
        \SetKwInOut{KwRequire}{Require}
        \SetKwInOut{KwOutput}{Output}
    
        \KwRequire{Frame window size \( w \), evaluation mode \( M \), $\ourpivot$ model \( \pivotf \), vision language model \(\vlm\), semantic score mapping function $f_{\text{VLM}}(\cdot)$, video automaton generation function $\xi(\cdot)$ probabilistic model checking function $\Psi(\cdot)$, probability mapping function $\ecdf(\cdot)$, probability distribution $\mathcal{D}$}
        \KwInput{Text prompt \( T \), text-to-video model \( \ttv \)}
        \KwOutput{$\neusv$ score $S_{\neusv}$}
        
        \Begin{
            $S \leftarrow \{\}$ \tcp*{Initialize an empty set for $\neusv$ scores of each evaluation mode}
            
            $\mathcal{V} \leftarrow \ttv(T)$ \tcp*{Generate a  video}
            
            \For{$m \in M$}{
            
            \For{$n = 0$ \KwTo $\mathrm{length}(\mathcal{V}) - w$ \KwStep $w$}{
                $C^\star \leftarrow \{\}$ \tcp*{Initialize an empty set for semantic confidence}
                
                $F \leftarrow \{ \mathcal{V}[n], \mathcal{V}[n+1], \dots, \mathcal{V}[n+w-1] \}$ \tcp*{Select a sequence of frames}
            
                $\propset, \ \spec \leftarrow \pivotf(T, m)$ \tcp*{Translate the T to  $\propset$ and $\spec$ for $m$}
                
                \For{$\prop_i \in \propset$}{
                    
                    $c_i \leftarrow \vlm(p_i, \mathcal{F})$ \tcp*{Obtain semantic confidence scores}

                    $c^\star_i \leftarrow f_{\text{VLM}}(c_i;\gamma_{fp})$ \tcp*{Calibrate $c_i$ with false positive threshold $\gamma_{fp}$}
                    
                    $C^\star[p_{i,n}] \leftarrow c^\star_i$ \tcp*{Append $c^\star_i$ to the semantic confidence set}
                    
                    \EndFor
                    }
                \EndFor
                }
                $\va \leftarrow \xi(\propset,C^\star)$ \tcp*{Construct $\va$}
                
                $\probsat = \Psi(\va, \spec)$ \tcp*{Obtain satisfaction probability}

                $s_m = \ecdf(\probsat, D_m)$ \tcp*{Calculate the calibrated score}
                $S \leftarrow S \cup \{s_m\}$ \tcp*{Append the score $s_m$ to the set $S$}
            }
            \EndFor
        
        $S_{\neusv} \leftarrow \frac{\sum S}{\mid S \mid}$
        }
        \caption{NeuS-V}
        \label{alg:neus_v}
    \end{algorithm} 
}
\endgroup
}

    \subsection{Automaton Representation of Synthetic Video} 
        Next, given $C^\star$ across all frames and propositions in $\propset$, we construct the video automaton $\va$ to represent the synthetic video as a DTMC $\dtmc$. This representation is crucial, as it enables the evaluation of synthetic video to determine whether it satisfies the given $\spec$ corresponding to the original text prompt. We build $\va$ using an automaton generation function $\xi : \propset \times C^\star \to \va$, defined as

        \begin{equation} 
            \label{eq:va}
            \va = \xi(\propset, C^\star) = \dtmc = (Q,q_0,\delta,\lambda),
        \end{equation}

        where $Q=\{q_0 \dots q_t\}$  is the set of possible states, each $q \in Q$ represents a unique configuration of truth values for $\propset$ with $q_0$ denoting the initial state. Given $C^\star$ for all atomic propositions in each sequence of frames from the synthetic video, the transition function $\delta(q,q')$ is defined as

        \begin{equation}
            \delta(q,q') = \prod_{i=1}^{|\propset|}(C^\star_{i})^{\mathbf{1}_{\{q'_{i}= 1\}}}(1-C^\star_{i})^{\mathbf{1}_{\{q'_{i} = 0\}}},
        \end{equation}

        where $\mathbf{1}_{\{q'_{i} = 1\}}$ is an indicator function that takes the value 1 if $\prop_i$ is true in $q'$, and 0 otherwise. Similarly, $\mathbf{1}_{\{q'_{i} = 0\}}$ takes the value 1 if and only if $\prop_i$ is false in $q'$. This is obtained by a labeling function $\lambda(q)$ that maps each state $q$ to the boolean value of its proposition. Further details on the $\va$ construction process are in the Appendix.

    \subsection{Verifying Synthetic Video Formally}
        Given $\va$, we compute the satisfaction probability $\probsat$ by formally verifying $\va$ against $\spec$. Specifically, we use the model checking method STORM \citep{storm,stormpy} that utilizes a probabilistic computation tree logic (PCTL) variant of temporal logic to calculate the satisfaction probability of $\va$ with respect to $\spec$. This probability is defined as $\probsat = \Psi(\va, \spec)$, where $\Psi(\cdot)$ is the probabilistic model checking function. This is performed by analyzing the probabilistic transitions and state labels within $\va$.

        \paragraph{Final Score Calibration.}
            Lastly, we calibrate the satisfaction probability to a $\neusv$ score using a satisfaction probability mapping function based on its empirical cumulative distribution, defined as 
            
            \begin{equation} 
                S = \{s_1,\dots,s_m\} = \ecdf(\probsat, D_m) \ \forall  \ m \in M,
            \end{equation}
            
            where $D_m$ is the distribution of satisfaction probabilities of each evaluation mode from wide samples of synthetic videos, and $S$ is the set of each evaluation mode score $s_m$. We take the average of these scores to comprehensively capture the variety of temporal specifications in the prompts 
        
            \begin{equation}
                S_{\neusv} \leftarrow \frac{\sum S}{\mid S \mid}.
            \end{equation}

\section{Experimental Setup}
\label{sec:experiment}

    \ourpivot{} uses GPT-4o and o1-preview for prompt to TL translation, whereas \neusv{} relies on InternVL2-8B \citep{jain2024peekaboointeractivevideogeneration} as its choice of VLM for automaton construction. We use a context of three frames when prompting the VLM. All the system prompts and example outputs can be found in our Appendix. We also use GPT-4o for prompt suite construction.

    \paragraph{T2V Models.}
        We evaluate the performance of both closed-source and open-source text-to-video models using two distinct prompt sets. For closed-source models, we select Gen-3\footnote{\href{https://runwayml.com/research/introducing-gen-3-alpha}{https://runwayml.com/research/introducing-gen-3-alpha}} and Pika\footnote{\href{https://pika.art}{https://pika.art}}, and we use the open-source T2V-Turbo-v2\citep{li2024t2v} and CogVideoX-5B\citep{yang2024cogvideox}. We run them with default hyperparameters on 8$\times$ A100 GPUs.
    
    \paragraph{Prompts and Evaluation Suite.} 
        As we observe that existing benchmarks for T2V evaluation \citep{huang2024vbench, liu2024fetv, Ji_2024_CVPR} lack explicit temporal instructions, we present the \neusv{} prompt suite with a total of 360 prompts. These are designed to rigorously evaluate models on their ability to maintain temporal coherence and accurately follow event sequences. The suite spans four themes (``Nature", ``Human \& Animal Activities", ``Object Interactions", and ``Driving Data") across three complexity levels (basic, intermediate, and advanced) based on the number of temporal and logical operators. 
    
        To ensure an unbiased baseline, we conduct a human study instructing annotators to explicitly \textit{disentangle} visual quality from text-to-video alignment. Both our prompt suite and the results of the human evaluation study will be made publicly available upon acceptance. Further details are provided in the supplementary material.

\section{Results}
    Building on our experimental setup, we now turn to the empirical evaluation of \neusv. Our experiments are designed to address three central questions that motivate the necessity of our methodology.

    \begin{enumerate}
        \item Does \neusv{}’s focus on temporal fidelity translate to a higher correlation with human annotations compared to baselines that emphasize visual quality?

        \item To what extent does grounding in temporal logic and formal verification provide a more robust evaluation framework than approaches based solely on VLMs?

        \item Can the reliability of \neusv{} extend beyond our synthetic prompt suite to established larger-scale datasets?
    \end{enumerate}

    \subsection{Formal Evaluation of Text-to-Video Models}
        As outlined in \Cref{sec:experiment}, we benchmark both closed-source and open-source state-of-the-art models: Gen-3 and Pika (closed-source), alongside T2V-Turbo-v2 \citep{li2024t2v1, li2024t2v} and CogVideoX-5B \citep{yang2024cogvideox} (open-source). 

\begin{figure}[t]
            \centering
            \begin{subfigure}{0.48\linewidth}
                \includegraphics[width=\linewidth, trim=0 0 0 317, clip]{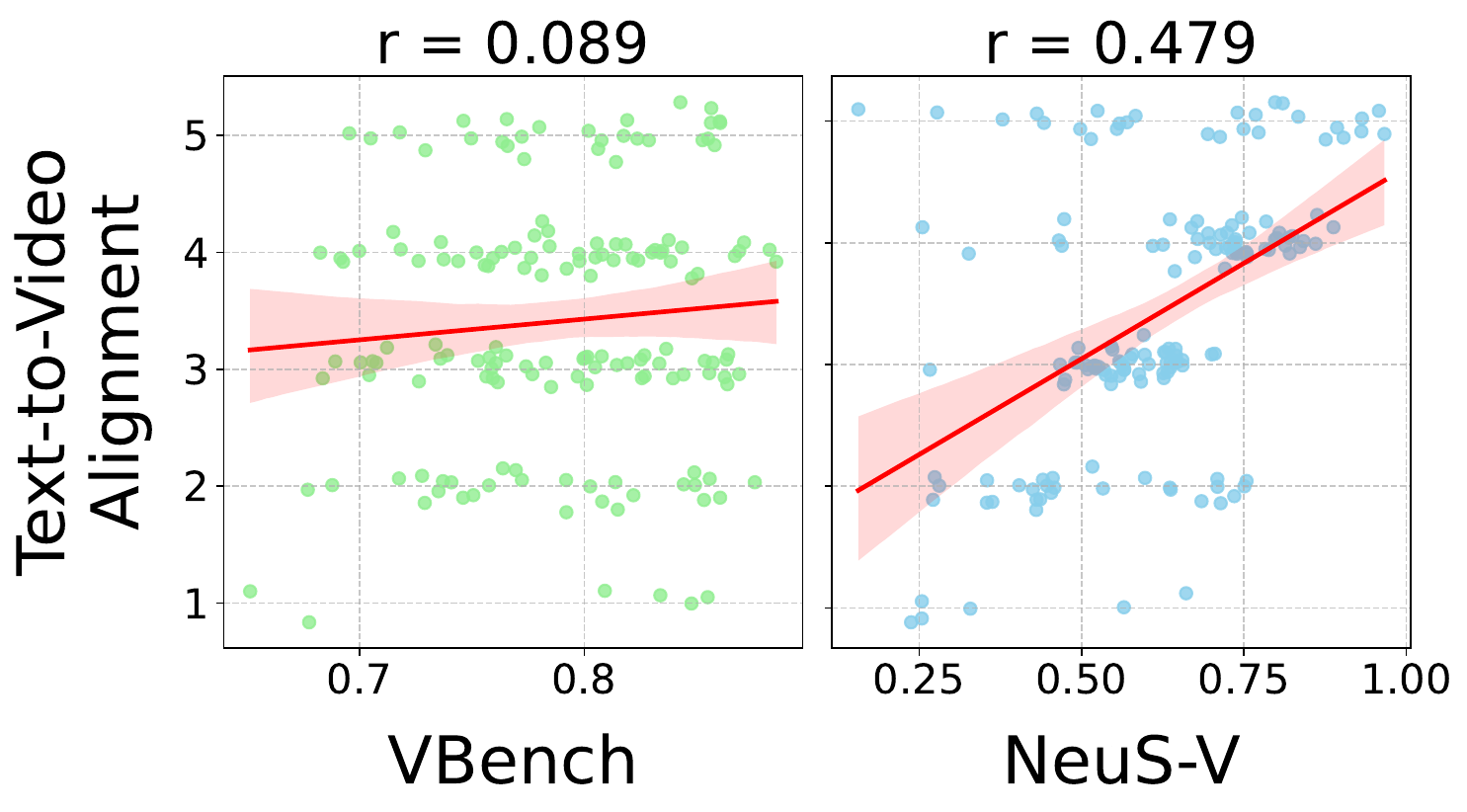}
                \caption{Gen-3}
            \end{subfigure} \hfill
            \begin{subfigure}{0.48\linewidth}
                \includegraphics[width=\linewidth, trim=0 0 0 317, clip]{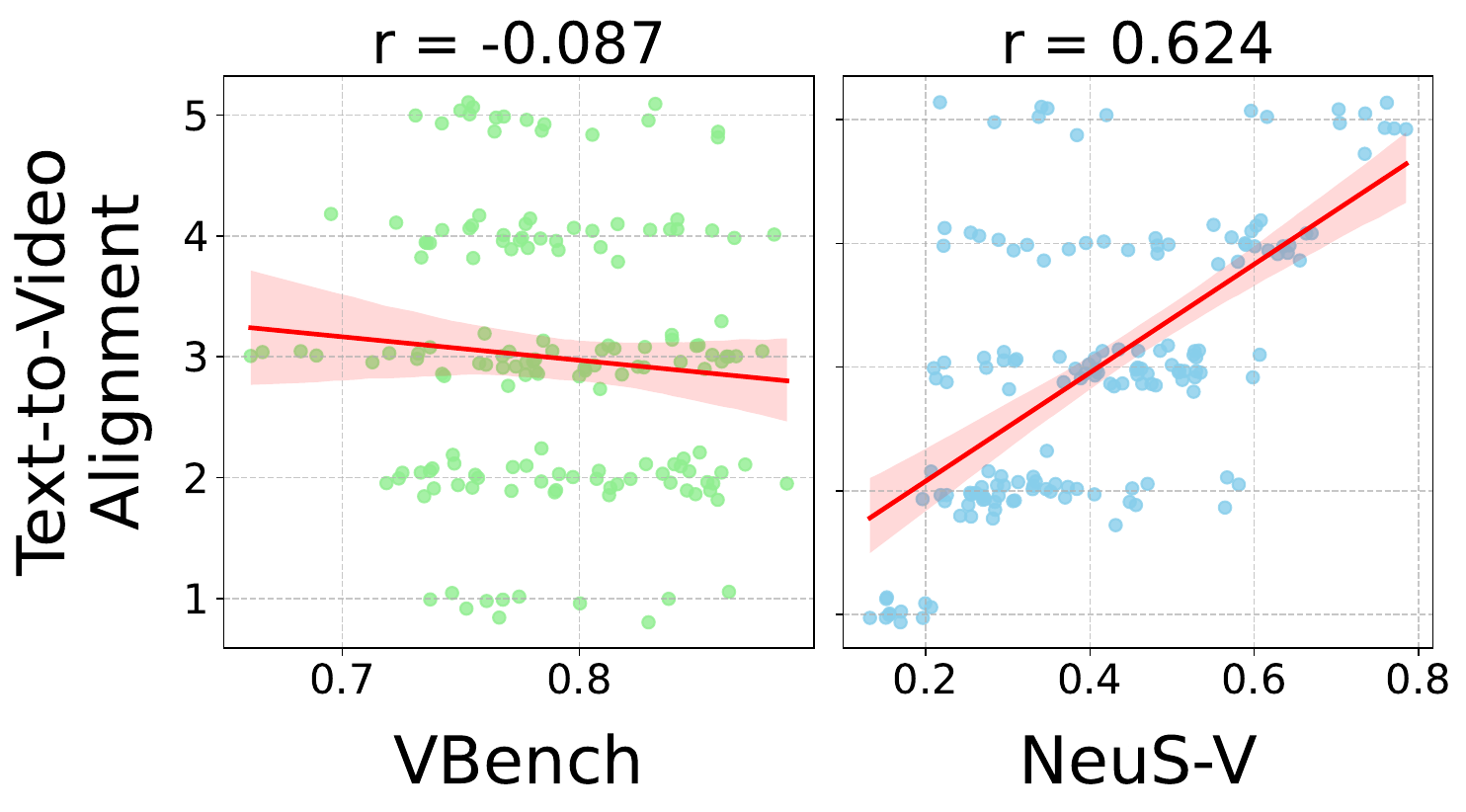}
                \caption{Pika}
            \end{subfigure}
            \par\smallskip
            \begin{subfigure}{0.48\linewidth}
                \includegraphics[width=\linewidth, trim=0 0 0 317, clip]{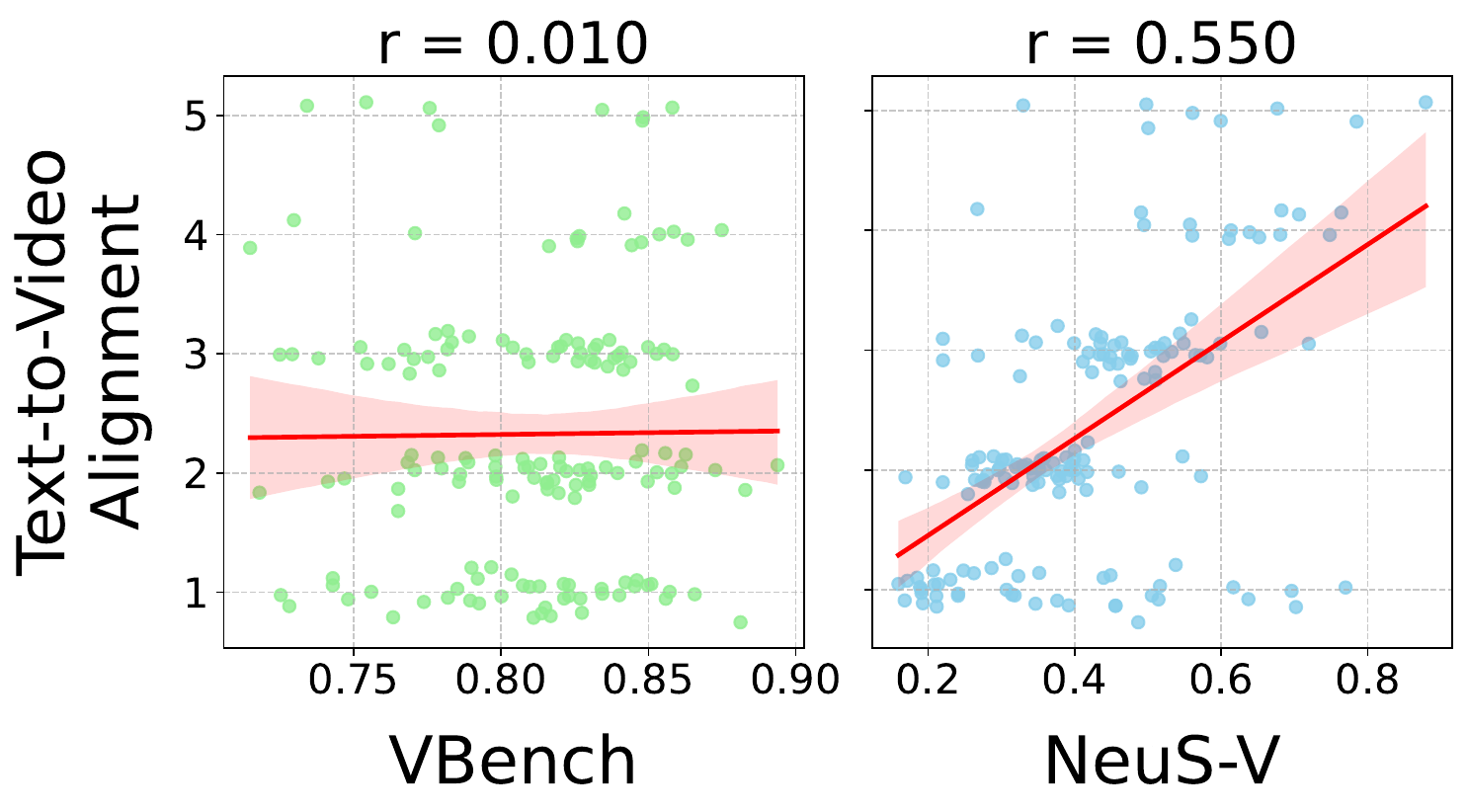}
                \caption{T2V-Turbo-v2\citep{li2024t2v}}
            \end{subfigure} \hfill
            \begin{subfigure}{0.48\linewidth}
                \includegraphics[width=\linewidth, trim=0 0 0 317, clip]{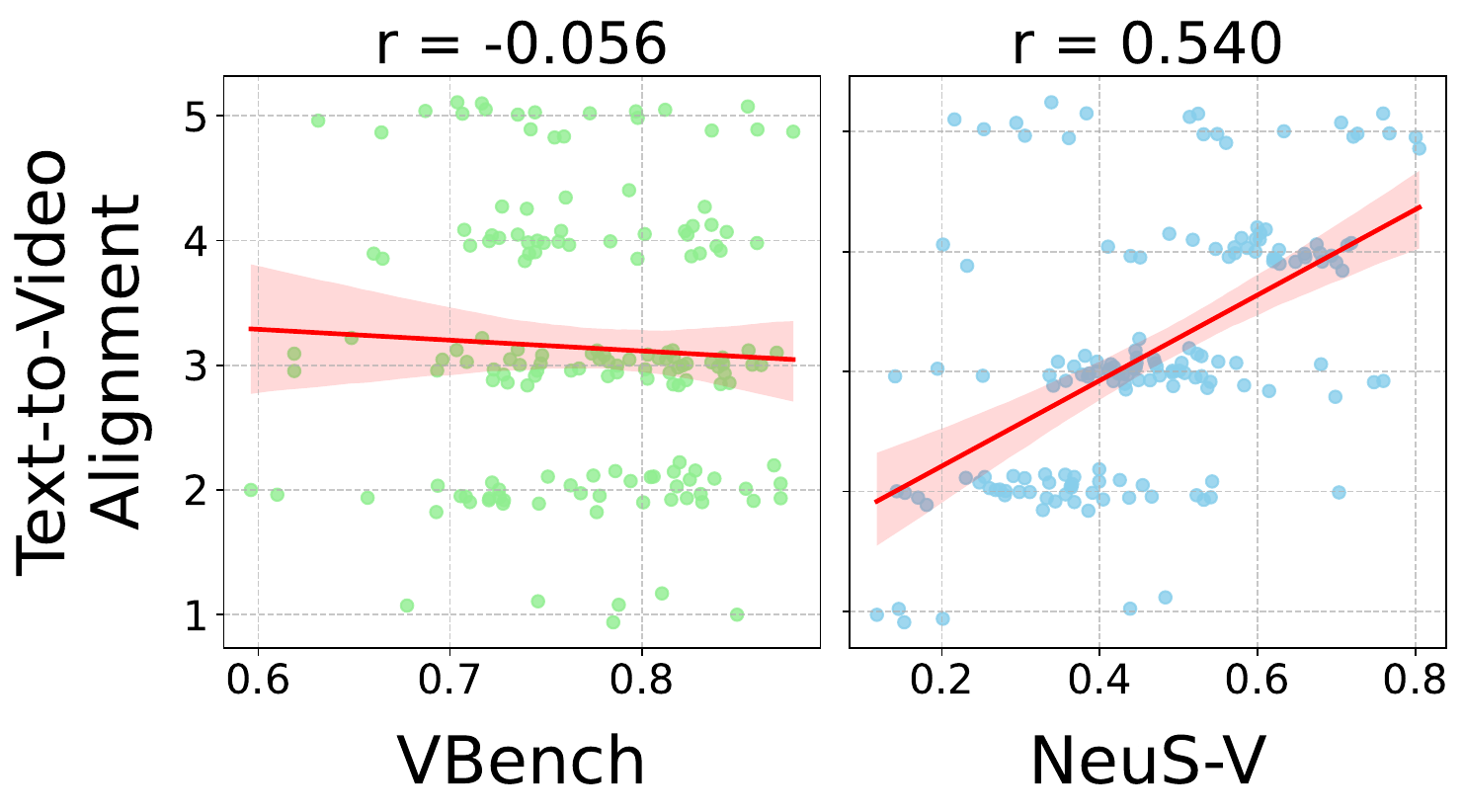}
                \caption{CogVideoX-5B\citep{yang2024cogvideox}}
            \end{subfigure}            
            \caption{\textbf{Correlation with Human Annotations.} \neusv{} consistently shows a stronger alignment with human text-to-video annotations (Pearson coefficients displayed at the top of each plot).}
            \label{fig:human-corr-vbench-neusv}
        \end{figure}

        \paragraph{Correlation with Human Annotations.} 
            We evaluate the selected video generation models using both \neusv{} and VBench, a well-established benchmark focused on visual quality. To assess correlation with human scores for text-to-video alignment, we plot these metrics in \Cref{fig:human-corr-vbench-neusv}. Across all models, \neusv{} demonstrates a consistently stronger correlation with human annotations, as indicated by the best-fit line and Pearson coefficient, outperforming VBench. By incorporating temporal logic specification and the automaton representation of synthetic videos, \neusv{} offers rigorous evaluations of text-to-video alignment. 
        
\begin{figure}[ht]
            \centering
            \begin{subfigure}{0.48\linewidth}
                \centering
                \includegraphics[width=\linewidth]{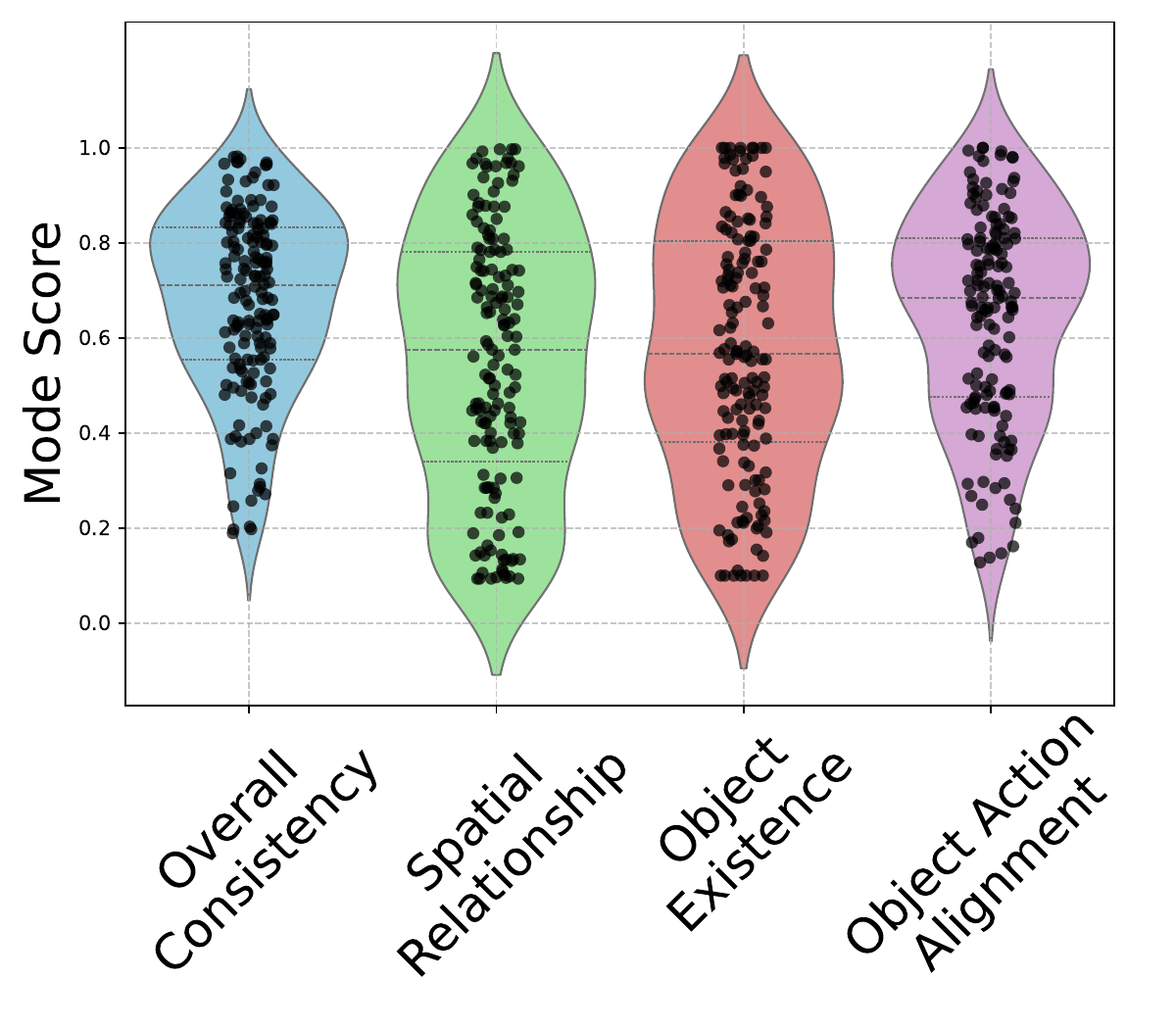}
                \caption{Gen-3}
            \end{subfigure}
            \begin{subfigure}{0.48\linewidth}
                \centering
                \includegraphics[width=\linewidth]{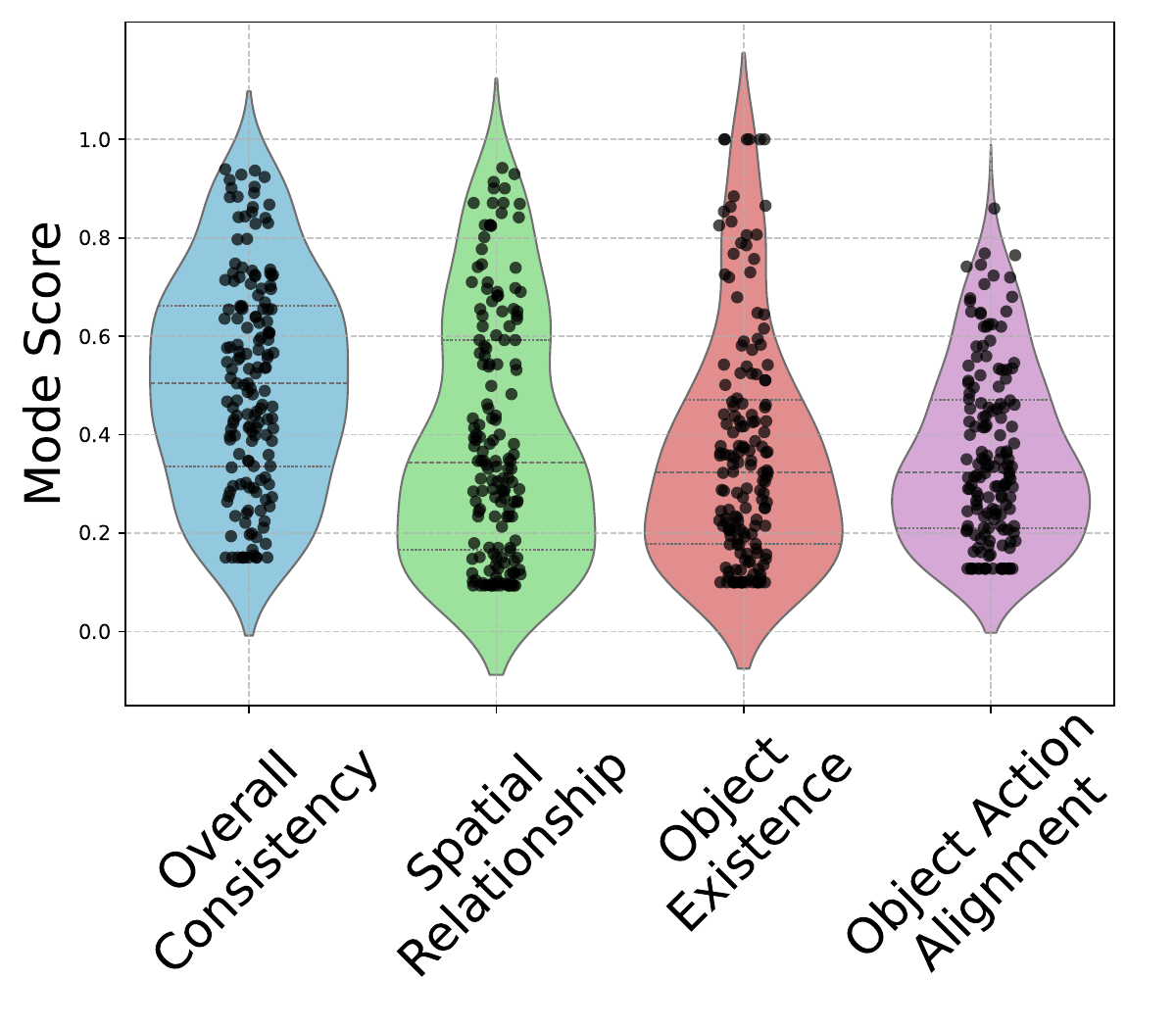}
                \caption{Pika}
            \end{subfigure}
            \begin{subfigure}{0.48\linewidth}
                \centering
                \includegraphics[width=\linewidth]{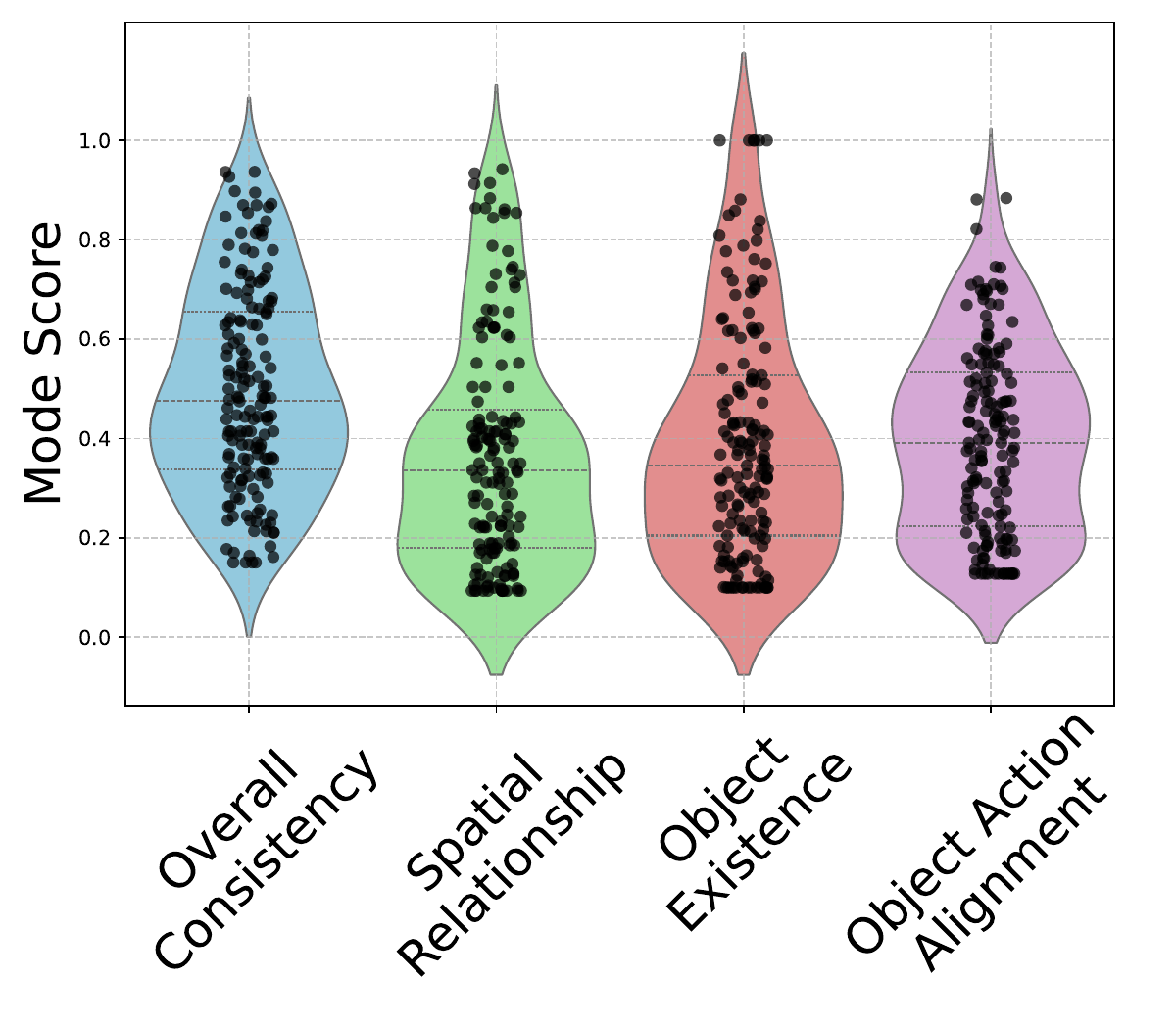}
                \caption{T2V-Turbo-v2}
            \end{subfigure}
            \begin{subfigure}{0.48\linewidth}
                \centering
                \includegraphics[width=\linewidth]{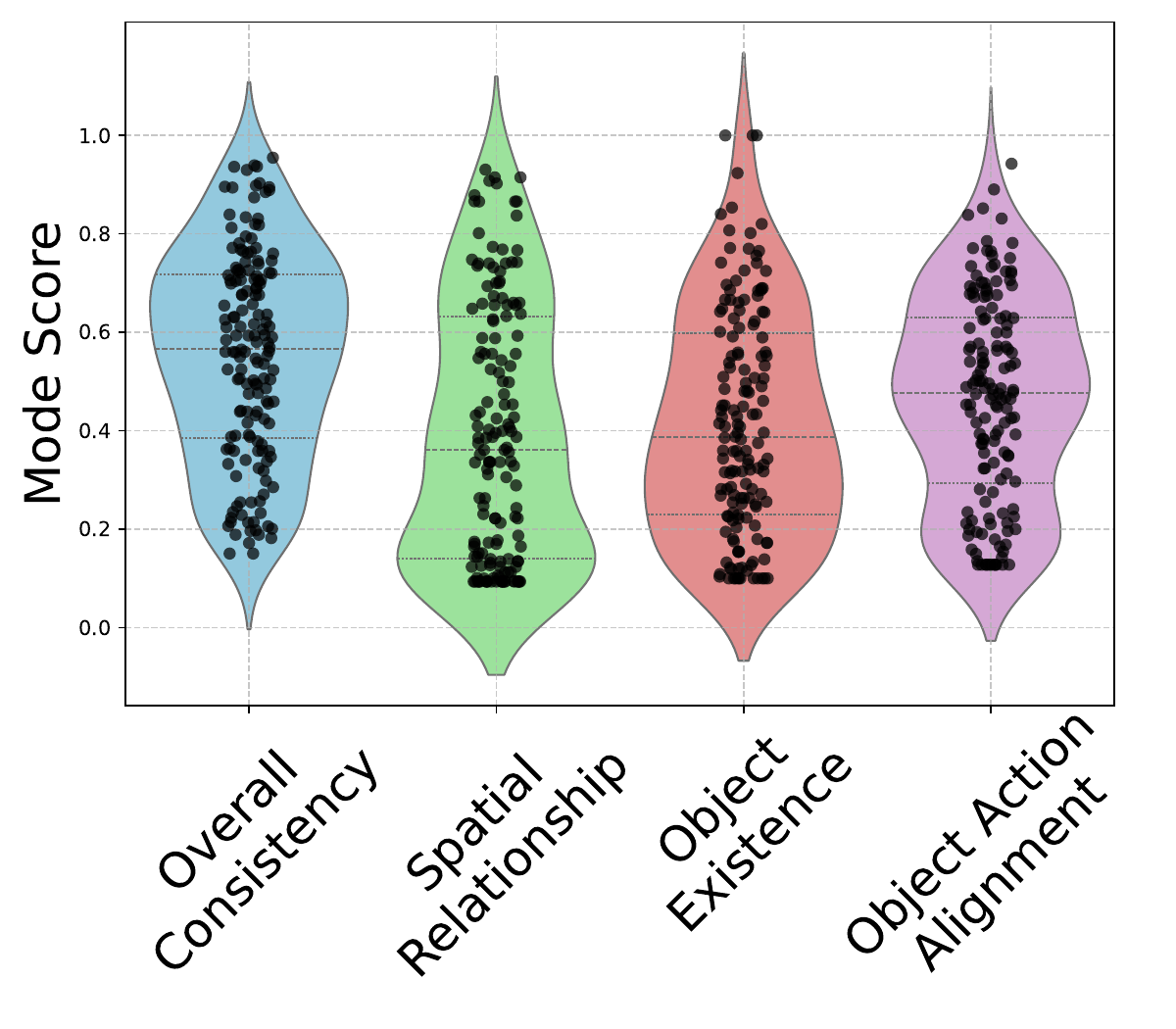}
                \caption{CogVideoX-5B}
            \end{subfigure}
            \caption{\textbf{Distribution of scores across the four modes of $\ourpivot$}}
            \label{fig:modes-of-temporal-logic}
        \end{figure}

        \paragraph{Performance on our Prompt Suite.} 
            We break down and analyze the performance of each T2V model on our evaluation suite, as shown in \Cref{tab:results-our-dataset}, organized by theme and complexity for a subset of 160 prompts. Results indicate that most models perform best when generating videos on the themes ``Nature" and ``Human and Animal Activities". The Gen-3 model achieves the highest scores across the suite, a result corroborated by human annotators. Furthermore, \Cref{fig:modes-of-temporal-logic} illustrates how each of the four evaluation modes in \ourpivot{} are utilized between models. Different prompts rely on distinct modes, indicating that a combination of these modes comprehensively captures the variety of temporal specifications in the prompts.

        \begin{table*}[t]
            \centering
            \resizebox{0.85\linewidth}{!}{
                \begin{tabular}{cccccc}
                    \toprule
                    \multicolumn{2}{c}{Prompts} & Gen-3 & Pika & T2V-Turbo-v2\citep{li2024t2v} & CogVideoX-5B\citep{yang2024cogvideox} \\
                    \midrule
                \multirow{4}{*}{By Theme} & Nature & 0.716 \textcolor{lightgray}{(0.47)} & 0.479 \textcolor{lightgray}{(0.70)} & 0.564 \textcolor{lightgray}{(0.46)} & 0.580 \textcolor{lightgray}{(0.53)} \\
                                               & Human \& Animal Activities & 0.752 \textcolor{lightgray}{(0.80)} & 0.531 \textcolor{lightgray}{(0.67)} & 0.564 \textcolor{lightgray}{(0.66)} & 0.623 \textcolor{lightgray}{(0.43)} \\
                                               & Object Interactions & 0.710 \textcolor{lightgray}{(0.16)} & 0.500 \textcolor{lightgray}{(0.40)} & 0.553 \textcolor{lightgray}{(0.66)} & 0.573 \textcolor{lightgray}{(0.65)} \\
                                               & Driving Data & 0.716 \textcolor{lightgray}{(0.48)} & 0.525 \textcolor{lightgray}{(0.66)} & 0.525 \textcolor{lightgray}{(0.30)} & 0.580 \textcolor{lightgray}{(0.52)} \\
                    \midrule
                    \multirow{3}{*}{By Complexity} & Basic (1 TL op.) & 0.774 \textcolor{lightgray}{(0.60)} & 0.589 \textcolor{lightgray}{(0.70)} & 0.610 \textcolor{lightgray}{(0.58)} & 0.641 \textcolor{lightgray}{(0.65)} \\
                                                   & Intermediate (2 TL ops.) & 0.680 \textcolor{lightgray}{(0.27)} & 0.464 \textcolor{lightgray}{(0.44)} & 0.508 \textcolor{lightgray}{(0.38)} & 0.549 \textcolor{lightgray}{(0.28)} \\
                                                   & Advanced (3 TL ops.) & 0.692 \textcolor{lightgray}{(-0.01)} & 0.400 \textcolor{lightgray}{(0.33)} & 0.494 \textcolor{lightgray}{(0.42)} & 0.550 \textcolor{lightgray}{(0.78)} \\
                    \midrule
                    \multicolumn{2}{c}{Overall Score} & 0.723 \textcolor{lightgray}{(0.48)} & 0.508 \textcolor{lightgray}{(0.62)} & 0.552 \textcolor{lightgray}{(0.55)} & 0.589 \textcolor{lightgray}{(0.54)} \\

                    \bottomrule
                \end{tabular}
            }
            \caption{\textbf{Benchmarking SOTA Text-to-Video Models.} Performance metrics reflect the full 360-prompt set, while correlations to human evaluations (in parentheses) are computed on a 160-prompt subset. \neusv{} enjoys high correlation across all themes and complexities.}
            \label{tab:results-our-dataset}
        \end{table*}

    \subsection{Ablation on Temporal Logic and Verification -- How Important is Formal Language?}
    \label{sec:ablation-on-tl-and-automata}
        In this ablation study, we investigate the role of formal grounding in temporal logic and model checking by replacing these components with VLM-based alternatives that lack such rigor. Specifically, instead of using \ourpivot{} to translate prompts into atomic propositions and specifications, we directly query an LLM to break down prompts into a set of yes/no questions covering the content and visual characteristics of each prompt. Next, rather than verifying with \neusv{}, we employ a VLM to score each frame independently based on the earlier generated questions. This ablation also reflects existing prompt-to-VQA baselines \citep{kou2024subjectivealigneddatasetmetrictexttovideo, liu2024evalcrafterbenchmarkingevaluatinglarge, zhang2024benchmarkingaigcvideoquality, Li_2024_CVPR}.

\begin{figure}
            \centering
            \begin{subfigure}{0.48\linewidth}
                \centering
                \includegraphics[width=\linewidth]{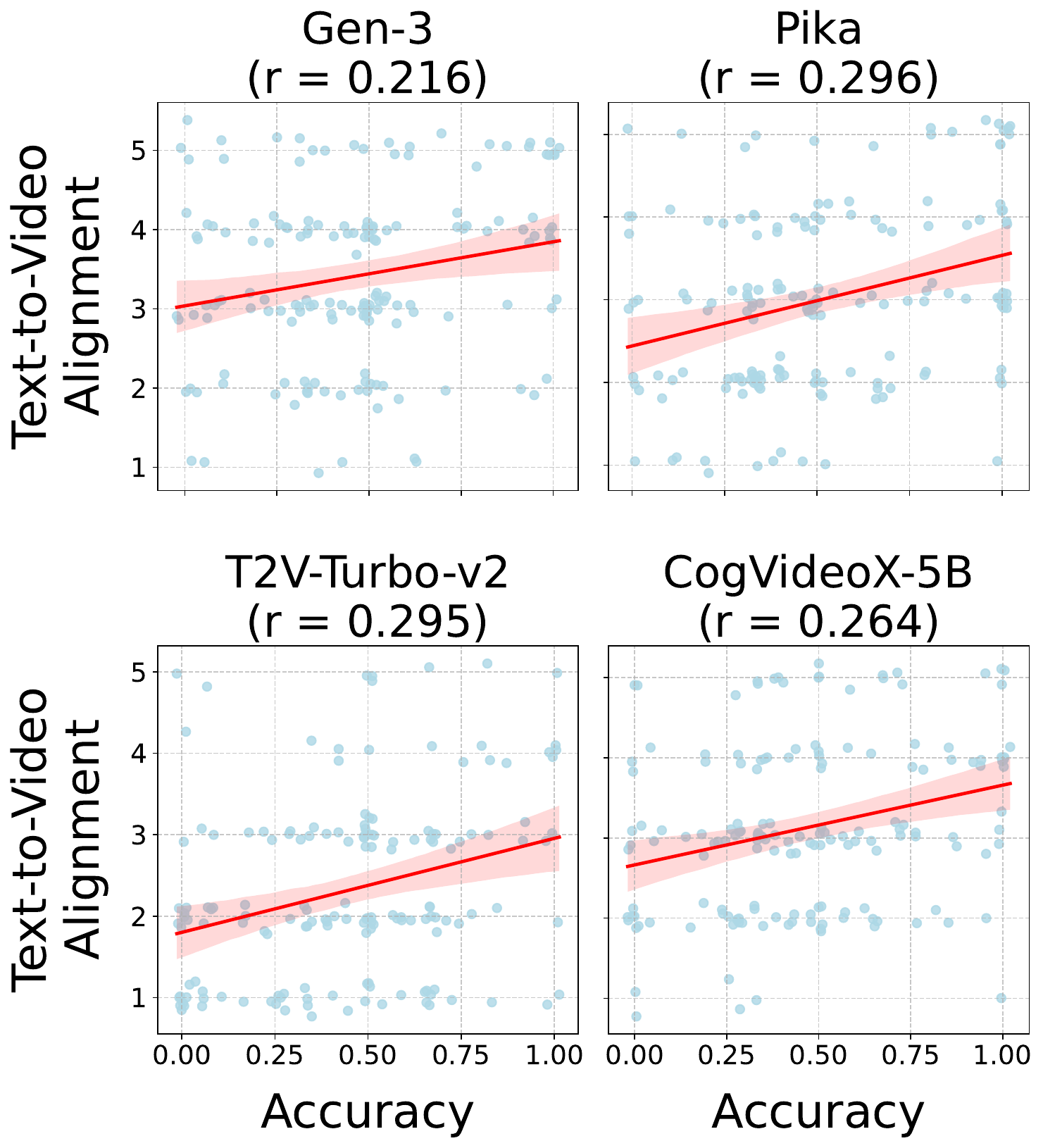}
                \caption{GPT-4o + LLaMa-3.2-11B}
            \end{subfigure}
            \hfill
            \begin{subfigure}{0.48\linewidth}
                \centering
                \includegraphics[width=\linewidth]{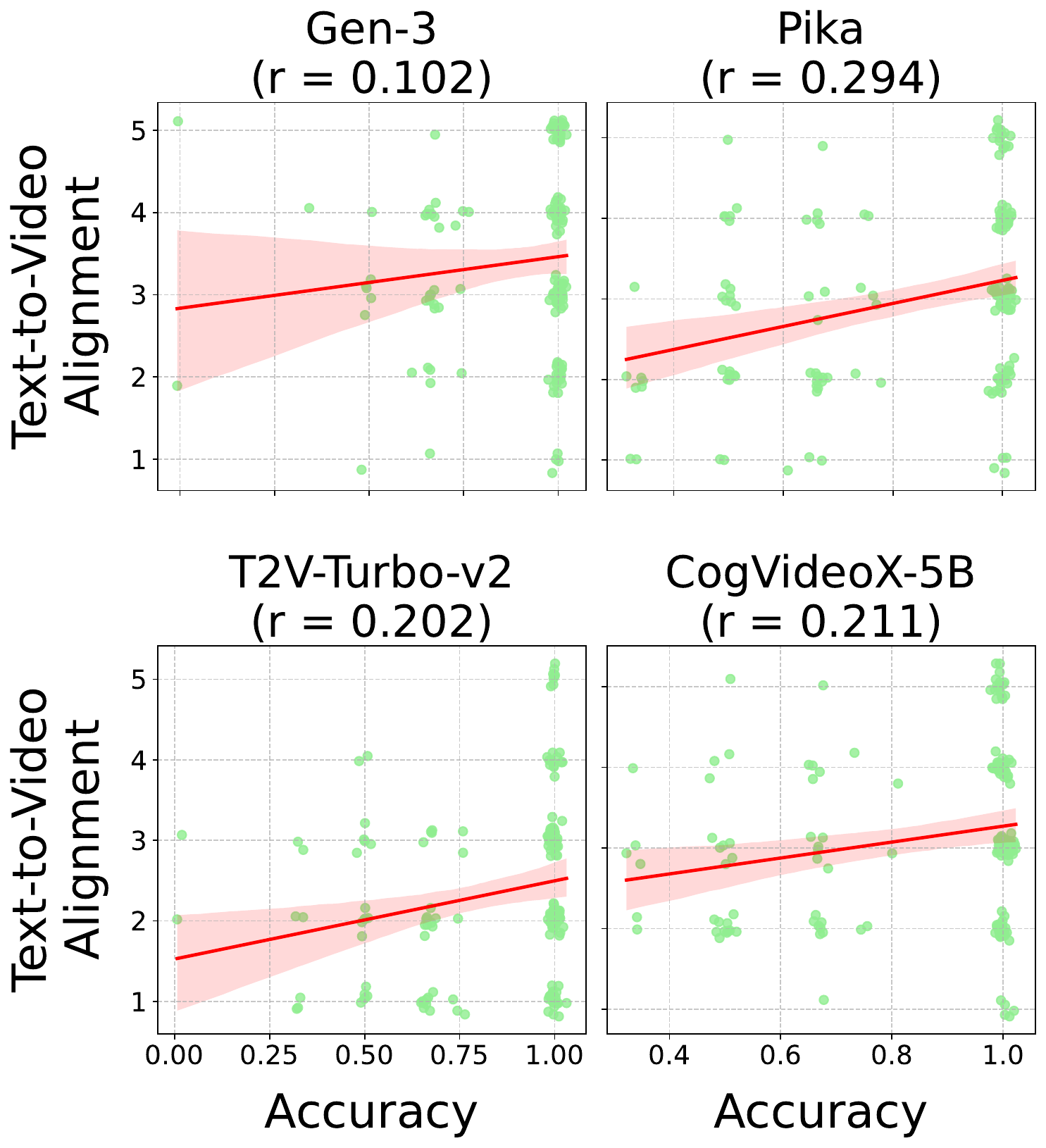}
                \caption{GPT-4o + LLaVA-Video-7B}
            \end{subfigure}
            \caption{\textbf{Is Formal Language Important?} VLMs without grounding in formal temporal logic lead to lower pearson coefficients (in brackets) as compared to \neusv{} from \Cref{fig:human-corr-vbench-neusv}.}
            \label{fig:ablation-on-tl-and-automata}
        \end{figure}

        For our setup, we use GPT-4o \citep{achiam2023gpt} as the LLM, along with two VLM configurations: LLaMa-3.2-11B-Vision-Instruct\footnote{\href{https://huggingface.co/meta-llama/Llama-3.2-11B-Vision}{https://huggingface.co/meta-llama/Llama-3.2-11B-Vision}} (capable of processing images) and LLaVA-Video-7B-Qwen2\footnote{\href{https://huggingface.co/lmms-lab/LLaVA-Video-7B-Qwen2}{https://huggingface.co/lmms-lab/LLaVA-Video-7B-Qwen2}} (capable of processing complete videos). We evaluate these methods with our prompt suite and the results are presented in \Cref{fig:ablation-on-tl-and-automata}. Our findings indicate that while both approaches show a positive correlation with human annotations (as indicated by Pearson’s coefficient), they do not achieve the same strength of correlation as the formal methods in \Cref{fig:human-corr-vbench-neusv} do. Also, LLaVA-Video exhibits overconfidence, resulting in a weaker correlation than the image-only LLaMa-3.2. These results underscore that the same VLMs when grounded in temporal logic and formal verification provide a more robust evaluation framework.

    \subsection{Demonstrating Robustness on a Real-World Video-Captioning Dataset}
        Our prompt evaluation suite, while effective, is modest in size with only 160 prompts. One might argue that this limited scale, along with the use of synthetic video captions, could raise doubts about the reliability of our metric. To address this concern, we use MSR-VTT \citep{xu2016msr}, a well-established dataset for video captioning -- a task which also involves assessing how well a video aligns with a given textual description. MSR-VTT features a large-scale set of video-caption pairs annotated by MTurk workers. We leverage their credibility and repurpose its validation set to evaluate the reliability of our metric. We create two splits: a positive set (where captions match videos) and a negative set (where captions do not match videos). We then test whether the \neusv{} metric can effectively distinguish between aligned and misaligned video-caption pairs.

        In \Cref{tab:msr-vtt-density}, we compare \neusv{} with VBench, reporting the mean and standard deviation of the scores for aligned and misaligned pairs. Our results show that \neusv{} successfully distinguishes between the positive and negative sets, assigning higher scores to aligned pairs and lower scores to misaligned ones. In contrast, VBench demonstrates a smaller difference between aligned and misaligned scores, highlighting \neusv{}'s superior ability to capture meaningful text-to-video alignment.

        \begin{table}[ht]
            \centering
            \resizebox{0.95\linewidth}{!}{
                \begin{tabular}{lccc}
                    \toprule
                    Metric & Aligned Captions & Misaligned Captions & Difference ($\uparrow$) \\
                    \midrule
                    VBench & 0.78 ($\pm$ 0.12) & 0.40 ($\pm$ 0.24) & 0.38 \\
                    \neusv & 0.82 ($\pm$ 0.10) & 0.30 ($\pm$ 0.18) & \textbf{0.52} \\
                    \bottomrule
                \end{tabular}
            }
            \caption{\textbf{Evaluating reliability on MSR-VTT.} \neusv{} distinguishes aligned and misaligned captions better than VBench.}
            \label{tab:msr-vtt-density}
        \end{table}

    \subsection{Ablations on Architectural Choices}
        \paragraph{Visual Context to VLM.}
            In this first ablation of architectural choices, we investigate whether using more frames necessarily improves the performance of \neusv. To measure this, we compare the use of a single frame versus three frames when prompting our VLM (InternVL2-8B). As shown in \Cref{tab:ablation-number-of-frames}, the use of three frames consistently results in a higher correlation with text-to-video alignment annotations. We did not test more than three frames, as this would exceed the context window of the LLM.

        \vspace{-1ex}
        \paragraph{Choice of VLM.}
            In the second ablation, we examine whether the choice of VLM affects automata construction. We test two recent VLMs: InternVL2-8B\citep{internvl2024v2} and LLaMa3.2-11B-Vision with single-frame contexts. As seen in \Cref{tab:ablation-choice-of-vlm}, InternVL2-8B consistently achieves higher correlation with text-to-video alignment annotations. LLaMa3.2 model tends to be overconfident in its outputs, despite attempts at calibration due to heavy skew in probabilities. This leads to a lower correlation with human labels.

            \begin{table}[t]
                \centering
                \resizebox{\linewidth}{!}{
                    \begin{tabular}{ccccc}
                        \toprule
                        VLM Context & Gen-3 & Pika & T2V-Turbo-v2 & CogVideoX-5B \\
                        \midrule
                        1 Frame & 0.859 \textcolor{lightgray}{(0.29)} & 0.613 \textcolor{lightgray}{(0.51)} & 0.590 \textcolor{lightgray}{(0.34)} & 0.715 \textcolor{lightgray}{(0.32)} \\
                        3 Frames & 0.614 \textcolor{lightgray}{(\textbf{0.48})} & 0.409 \textcolor{lightgray}{(\textbf{0.62})} & 0.417 \textcolor{lightgray}{(\textbf{0.55})} & 0.461 \textcolor{lightgray}{(\textbf{0.54})} \\
                        \bottomrule
                    \end{tabular}
                }
                \caption{\textbf{Impact of contextual frames to VLM.} More contextual information results in higher correlation with human annotations.}
                \label{tab:ablation-number-of-frames}
            \end{table}

            \begin{table}[t]
                \centering
                \resizebox{\linewidth}{!}{
                    \begin{tabular}{ccccc}
                        \toprule
                        Choice of VLM & Gen-3 & Pika & T2V-Turbo-v2 & CogVideoX-5B \\
                        \midrule
                        LLaMa3.2-11B & 0.921 \textcolor{lightgray}{(0.15)} & 0.660 \textcolor{lightgray}{(0.16)} & 0.730 \textcolor{lightgray}{(0.08)} & 0.785 \textcolor{lightgray}{(-0.02)} \\
                        InternVL2-8B & 0.859 \textcolor{lightgray}{(\textbf{0.29})} & 0.613 \textcolor{lightgray}{(\textbf{0.51})} & 0.590 \textcolor{lightgray}{(\textbf{0.34})} & 0.715 \textcolor{lightgray}{(\textbf{0.32})} \\
                        \bottomrule
                    \end{tabular}
                }
                \caption{\textbf{Impact of choice of VLM on correlation.} InternVL2-8B for automaton construction shows higher correlation with human annotations.}
                \label{tab:ablation-choice-of-vlm}
            \end{table}

\section{Discussion}
    \paragraph{Is Rigorous Formalism Key?}
        A natural question that arises is whether formal temporal logic can be replaced with simpler if-else style checks and assertions. In principle, yes. However, as prompt length and complexity increase, such rule-based approaches exhibit scaling challenges, making them impractical. In contrast, symbolic temporal operators offer a near-linear scaling to verify adherence to automata. As shown through \Cref{fig:ablation-on-tl-and-automata} in \Cref{sec:ablation-on-tl-and-automata}, even though such alternative approaches may show promise, they ultimately fall short of the robustness provided by formal temporal logic. \neusv{} is the first work that introduces formalism to T2V evaluation, going beyond quality-based metrics.

    \paragraph{Rethinking Text-to-Video Models.}
        Evaluations in \Cref{tab:results-our-dataset} reveal a significant limitation in current text-to-video models when handling temporally extended prompts. Although marketed as text-driven video generation, these models often fall short of creating motion and sequences that truly follow the given prompts. Much of the perceived ``motion" comes from simplistic effects like panning and zooming with minimal capacity for generating interactions or meaningful scene transitions. In practice, video-generation workflows require numerous rounds of re-prompting to achieve prompt adherence, whereas ideally, we should expect alignment with the prompt most times.

    \vspace{-1ex}
    \paragraph{Limitations and Future Work.}
        \neusv{} currently lacks a mechanism to penalize unintended elements in generated videos, making it difficult to distinguish between irrelevant artifacts and potentially useful auxiliary content that was not explicitly requested in the prompt. In the future, our approach could be used to refine the generated videos until they meet a set threshold, boosting prompt adherence without retraining. Furthermore, feedback from \neusv{} could also be used for prompt optimization or model training.

\section{Conclusion}
    Text-to-video models are increasingly being applied in safety-critical areas such as autonomous driving, education, and robotics, where maintaining temporal fidelity is essential. However, we find that most current evaluation methods prioritize visual quality over text-to-video alignment. To address this, we introduce \neusv{}, a novel metric that translates prompts into temporal logic specifications and represents videos as automata, thereby scoring them through formal verification. Alongside \neusv, we present a benchmark of temporally challenging prompts to assess the ability of state-of-the-art T2V models to follow specified event sequences. Our evaluation reveals limitations in these models, which frequently rely on simplistic camera movements rather than true temporal dynamics. We aim to pave the way for the development of future T2V models that achieve greater alignment with temporal requirements.

\section*{Acknowledgments}
\label{sec:acknowledgements}
This material is based upon work supported in part by the Office of Naval Research (ONR) under Grant No. N00014-22-1-2254. Additionally, this work was supported by the Defense Advanced Research Projects Agency (DARPA) contract DARPA ANSR: RTX CW2231110. Approved for Public Release, Distribution Unlimited.


{
    \small
    \bibliographystyle{ieeenat_fullname}
    \bibliography{references}
}


\clearpage
\setcounter{page}{1}
\maketitlesupplementary

\section{Temporal Logic Operation Example}
\label{sec:appx_temporal_logic}
Given a set of atomic propositions $\mathcal{P}=$ \{\text{Event A}, \text{Event B}\}, the TL specification $\Phi=$ \always~Event A (read as ``Always Event A") means that `Event A' is \texttt{True} for every step in the sequence. Additionally, $\Phi=$ \eventually~Event B (read as ``eventually event b") indicates that there exists at least one `Event B' in the sequence. Lastly, $\Phi=$ Event A \until \hspace{1pt} Event B (read as ``Event A Until Event B") means that `Event A' exists until `Event B' becomes \texttt{True}, and then `Event B' remains \texttt{True} for all future steps.

\section{Prompt Understanding via Temporal Logic
Specification (\ourpivot{})}
\label{sec:appx_puls}
In order to obtain $\mathfrak{D}_\text{train}$, we first begin with the larger dataset $\mathfrak{D}$ with size $\mathcal{B}$ where $\mathfrak{D} = \mathfrak{D}_\text{T2P} \cup \mathfrak{D}_\text{T2TL}$. $\mathfrak{D}_\text{T2P}$ and $\mathfrak{D}_\text{T2TL}$ are defined as the following:
\begin{equation} 
    \mathfrak{D}_\text{T2P} = \{ (\mathcal{T}_i, \propset_i) \}_{i=1}^{\mathcal{B}}, \quad \mathfrak{D}_\text{T2P} \subset \mathfrak{D},
\end{equation}
\begin{equation}
    \mathfrak{D}_\text{T2TL} = \{ (\mathcal{T}_i, \propset_i,\spec_i)  \}_{i=1}^{\mathcal{B}}, \quad \mathfrak{D}_\text{T2TL} \subset \mathfrak{D}.
\end{equation}
Using these datasets, we use \ourpivot{} to find a specification $\Phi$ for each mode and for a given text prompt $\mathcal{T}$ using the following algorithm.



    \begin{algorithm}[h]
        \footnotesize
        \DontPrintSemicolon
        \SetKwInOut{KwRequire}{Require}
        \KwRequire{LLM Prompt Optimizer \texttt{MIPROv2}}
        \KwInput{List of mode $M$, text 
        prompt $\mathcal{T}$, training examples $\mathfrak{D}_\text{T2P}$ and $\mathfrak{D}_\text{T2TL}$, number of few shot examples $\mathcal{N}$}
        \KwPrompt{Modules $\ttp$ and $\tttl$}
        \KwOutput{TL Specification $\Phi$}
        \Begin{
        $\Phi \leftarrow \{\}$ \tcp*{Initialize empty set $\Phi$}
            \For{$m \in M$}{
                $\dttp = \texttt{MIPROv2.optimize}\left(\mathfrak{D}_\text{T2P}, m, \mathcal{N}\right)$ \tcp*{Find optimal fewshot dataset}
                $\dttl = \texttt{MIPROv2.optimize}\left(\mathfrak{D}_\text{T2TL}, m, \mathcal{N}\right)$ \tcp*{Find optimal fewshot dataset}


                $\oprompt_\text{T2P} \leftarrow \{ \left(\mathcal{T}_{i}, \propset_i\right) \mid \mathcal{T}_i, \propset_i \in \dttp\}_{i=1}^{\mathcal{N}}$ \tcp*{Update parameters}
                $\oprompt_\text{T2TL} \leftarrow \{ \left(\mathcal{T}_{i}, \propset_i, \spec_i\right) \mid \mathcal{T}_i, \propset_i, \spec_i \in \dttl\}_{i=1}^{\mathcal{N}}$ \tcp*{Update parameters}

                $\mathcal{P} = \ttp \left(\mathcal{T}, m, \oprompt_{\text{T2P}}\right)$ \tcp*{Compute propositions}
                $\Phi \leftarrow \Phi \cup \{\tttl \left(\mathcal{T}, \mathcal{P}, m, \oprompt_{\text{T2TL}}\right)\}$ \tcp*{Compute specification}

            \EndFor
            }
            \Return $\Phi$
        }
        \caption{\ourpivot{}}
        \label{alg:pd-pseudocode}
    \end{algorithm} 

\subsection{DSPy \& MIPROv2}
To evaluate L4-5 of \Cref{alg:pd-pseudocode}, \ourpivot{} uses DSPy and MIPROv2 to optimize the prompts by selecting the appropriate subset of $\mathfrak{D}_\text{T2P}$ and $\mathfrak{D}_\text{T2TL}$ respectively. First, it creates a bootstrap dataset $\mathfrak{D}'$ from the original dataset $\mathfrak{D}$. This dataset comprises effective few-shot examples that are generated using rejection sampling. Since the bootstrapping process is done for both $\mathfrak{D}_\text{T2P}$ and $\mathfrak{D}_\text{T2TL}$, we can say $\mathfrak{D}' = \mathfrak{D}_\text{T2P}' \cup \mathfrak{D}_\text{T2TL}'$.

Next, \ourpivot{} proposes $k$ different instructions using an LLM depending on the properties of the original dataset $\mathfrak{D}$ and the original instruction, yielding the instruction sets $\mathbf{x}_\text{T2P} = \{x_{\text{T2P}_j}\}_{j=0}^k$ and $\mathbf{x}_\text{T2TL} = \{x_{\text{T2TL}_j}\}_{j=0}^k$. Thus, given a particular dataset entry $i$ and instruction $j$:
\begin{equation} 
    \propset_{i, j}^\text{pred} = \texttt{LLM}(\mathcal{T}_i, x_{\text{T2P}_j}, \theta_{llm}), \quad \mathcal{T}_i \in \mathfrak{D}_\text{T2P}'
\end{equation}
\begin{equation} 
    \spec_{i, j}^\text{pred} = \texttt{LLM}(\mathcal{T}_i, \propset_i, x_{\text{T2TL}_j}, \theta_{llm}), \quad \mathcal{T}_i, \propset_i \in \mathfrak{D}_\text{T2TL}'
\end{equation}

Accuracy functions are defined as the following, where $\pi$ is an evaluation metric that returns a similarity score between 0 and 1 of its inputs:
\begin{equation} 
    f_\text{T2P}(\mathcal{D}_\text{T2P}', i, j) = \pi(\mathcal{P}_i, \mathcal{P}_{i, j}^{pred}), \quad \propset_i \in \mathfrak{D}_\text{T2P}'
\end{equation}
\begin{equation} 
    f_\text{T2TL}(\mathcal{D}_\text{T2TL}', i, j) = \pi(\Phi_i, \Phi_{i, j}^{pred}), \quad \Phi_i \in \mathfrak{D}_\text{T2TL}'
\end{equation}

Bayesian Optimization is used to create an optimal subset, $\mathfrak{D}_\text{train}$ of size $\mathcal{N}$, with the following operation:
\begin{equation}
\begin{split}
    \dttp = &\arg\max_{\mathfrak{D}_s \subset \mathfrak{D}_\text{T2P}', |\mathfrak{D}_s| = \mathcal{N}} \\ &\left[\sum_{i \in \mathfrak{D}_s} \max_{0 \leq j \leq k} f_\text{T2P}(\mathfrak{D}_s, i, j)\right]
\end{split}
\end{equation}
\begin{equation}
\begin{split}
    \dttl &= \arg\max_{\mathfrak{D}_s \subset \mathfrak{D}_\text{T2TL}', |\mathfrak{D}_s| = \mathcal{N}} \\ &\left[\sum_{i \in \mathfrak{D}_s} \max_{0 \leq j \leq k} f_\text{T2TL}(\mathfrak{D}_s, i, j)\right]
\end{split}
\end{equation}


\begin{equation}
    \mathfrak{D}_\text{train} = \dttp \cup \dttl.
\end{equation}

With this process, $\mathfrak{D}_\text{train}$ can be used as an effective few-shot dataset for \ourpivot{}. $\ttp$ uses $\dttp$ and $\tttl$ uses $\dttl$ as their datasets. \Cref{prompt:t2p} and \Cref{prompt:t2tl} detail the system prompts for the overall consistency mode in $\ttp$ and $\tttl$ respectively.

\begin{minipage}{.95\linewidth}
    \begin{tcolorbox}[
        title=\ourpivot{} Prompt for Text to Propositions (.md),
        colback=white,
        colframe=gray,
        colbacktitle=gray
        ]
        \ttfamily\scriptsize
        **System Message**:

\smallskip
Your input fields are: \\
1. `input\_prompt' (str): Input prompt summarizing what happened in a video.

\smallskip
Your output fields are: \\
1. `reasoning' (str) \\
2. `output\_propositions' (str): A list of atomic propositions that correlate with the inputted prompt. For example, for a prompt such as `A person holding a hotdog is  walking down to the street where many cars next to the huge truck', the propositions are `person holds hotdog', `person walks', and `car next to truck'. This outputted list of propositions MUST be formatted as: [prop1, prop2, prop3].

\smallskip
Your objective is: \\
        Convert from a prompt to a list of propositions using the following schema. \\
\smallskip
\hrule height 0.05pt
\medskip

**User message**:

\smallskip
\underline{Input Prompt}: A boat sailing leisurely along the Seine River with the Eiffel Tower in background, zoom out.

\smallskip
Respond with the corresponding output fields, starting with the field `reasoning', then `output\_propositions'.

\bigskip
**Assistant message**:

\smallskip
\underline{Reasoning}: Not supplied for this particular example.

\smallskip
\underline{Output Propositions}: [`There is a boat', `The boat is sailing leisurely', 'The boat is along the Seine River', `The Eiffel Tower is in the background', `The view is zooming out']

\bigskip
\hspace{7em}$\cdots$ \\
\smallskip
\textbf{< redacted more examples >} \\
\smallskip
\hspace{7em}$\cdots$
    \end{tcolorbox}
    \captionsetup{type=prompt}
    \captionof{prompt}{\textbf{T2P Prompt for \ourpivot.} System prompt to map prompts to the propositions for the ``Overall Consistency" mode. Other modes follow a similar structure, albeit with minor changes in instructions and few-shot examples.}
    \label{prompt:t2p}
\end{minipage}

\section{Vision Language Model Calibration}
\label{sec:appx_vlm_calibration}
In this section, we provide the implementation details to detect the existence of propositions obtained by \ourpivot{} to label each frame in the synthetic video. First, we describe how VLMs conduct inference and then describe the methodology for obtaining the optimal thresholds for calibrating the vision language model.

\subsection{Inference Via Vision Language Models}
We use VLM as a semantic detector as explained in \Cref{subsec:semantic_score_from_neural_perception_model}. We pass each atomic proposition $p_i \in \mathcal{P}$ such as ``person", ``car", ``person in the car", etc. along with \Cref{prompt:vlm}. Once the VLM outputs either `Yes' or `No', we compute the token probability of the response and use it as a confidence score for the detection.

\begin{minipage}{.95\linewidth}
    \begin{tcolorbox}[
        title=\ourpivot{} Prompt for Text to Temporal Logic (.md),
        colback=white,
        colframe=gray,
        colbacktitle=gray
        ]
        \ttfamily\scriptsize
        **System Message**:

\smallskip
Your input fields are: \\
1. `input\_prompt` (str): Input prompt summarizing what happened in a video. \\
2. `input\_propositions` (str): A list of atomic propositions in the video.

\smallskip
Your output fields are: \\
1. `reasoning' (str) \\
2. `output\_specification' (str): The formal specification of the event sequence. The response will be a linear temporal logic (LTL) formula made of the inputted propositions combined with LTL symbols. Each symbol can only be one of [`AND', `OR', `NOT', `UNTIL', `ALWAYS', `EVENTUALLY'].

\smallskip
Your objective is: \\
        Convert from a prompt and a list of propositions to a TL specification using the following schema. \\
\smallskip
\hrule height 0.05pt
\medskip

**User message**:

\smallskip
\underline{Input Prompt}: A baseball glove on the right of a tennis racket, front view.

\smallskip
\underline{Input Propositions}: [`There is a baseball glove', `There is a tennis racket', `The baseball glove is on the right of the tennis racket', `The view is from the front']

\smallskip
Respond with the corresponding output fields, starting with the field `reasoning', then `output\_specification'.

\bigskip
**Assistant message**:

\smallskip
\underline{Reasoning}: Not supplied for this particular example.

\smallskip
\underline{Output Specification}: (There is a baseball glove AND There is a tennis racket AND The baseball glove is on the right of the tennis racket AND The view is from the front)

\bigskip
\hspace{7em}$\cdots$ \\
\smallskip
\textbf{< redacted more examples >} \\
\smallskip
\hspace{7em}$\cdots$
    \end{tcolorbox}
    \captionsetup{type=prompt}
    \captionof{prompt}{\textbf{T2TL Prompt for \ourpivot.} System prompt to map prompts and propositions for the ``Overall Consistency" mode. Other modes follow a similar structure, albeit with minor changes in instructions and few-shot examples.}
    \label{prompt:t2tl}
\end{minipage}

\begin{minipage}{0.95\linewidth}
    \begin{tcolorbox}[
        title=Prompt for Semantic Detector (VLM),
        colback=white,
        colframe=gray,
        colbacktitle=gray]
    \label{prompt:vlm}
        \ttfamily\scriptsize
        Is there \{atomic proposition ($\prop_i$)\} present in the sequence of frames?

        [PARSING RULE] 1. You must only return a Yes or No, and not both, to any question asked.
        
        2. You must not include any other symbols, information, text, or justification in your answer or repeat Yes or No multiple times.

        3. For example, if the question is 'Is there a cat present in the Image?', the answer must only be 'Yes' or 'No'.
    \end{tcolorbox}
    \captionsetup{type=prompt}
    \captionof{prompt}{\textbf{Semantic Detector VLM.} Used to identify the atomic proposition within the frame by initiating VLM with a single frame or a series of frames.}
    \label{prompt:semantic-detector-vlm}
\end{minipage}

\subsection{False Positive Threshold Identification}

\paragraph{Dataset for Calibration:}
We utilize the COCO Captions \cite{chen2015microsoft} dataset to calibrate the following open-source vision language models -- InternVL2  Series (1B, 2B, 8B) \cite{internvl2024v2} and LLaMA-3.2 Vision Instruct \cite{metallama32v} -- for \neusv{}. Given that each image-caption pair in the dataset is positive coupling, we construct a set of negative image-caption pairs by randomly pairing an image with any other caption corresponding to a different image in the dataset. Once we construct the calibration dataset, which comprises 40000 image caption pairs, we utilize the VLM to output a `Yes' or a `No' for each pair.

\paragraph{Thresholding Methodology}

\begin{figure}[t]
    \centering
    \includegraphics[width=\linewidth, trim=1in 0 1in 0, clip]{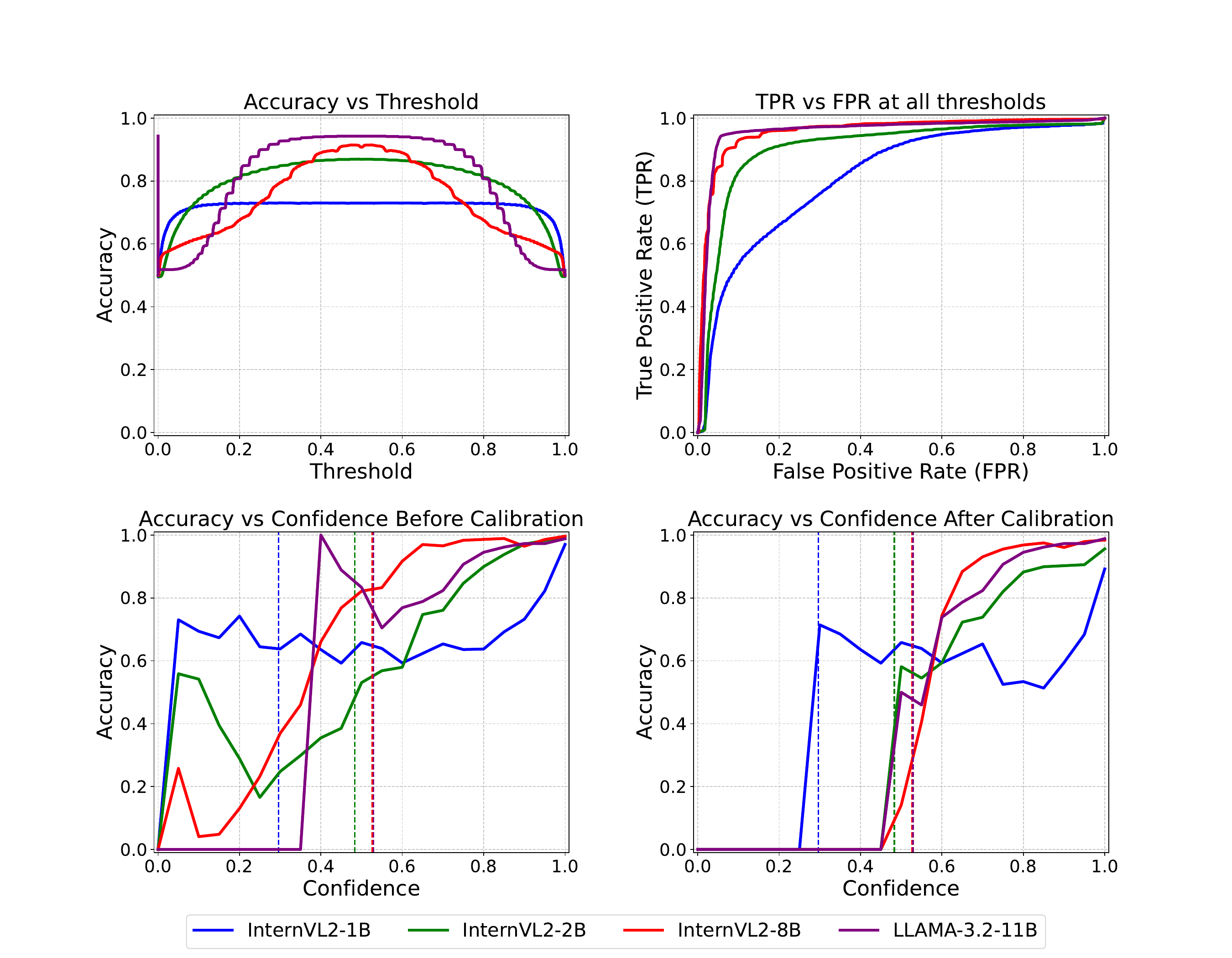}
    \caption{\textbf{Calibration Plots.} We plot the accuracy vs threshold for all VLMs on our calibration dataset constructed from the COCO Caption dataset (top left). We plot the True Positive Rate (TPR) vs False Positive Rate (FPR) across all thresholds on the top right. Finally, the bottom plots show the confidence vs accuracy of the model before and after calibration, respectively.}
    \label{fig:calibration}
\end{figure}

We can identify the optimal threshold for the VLM by treating the above problem as either a single-class or multi-class classification problem. We opt to do the latter. The process involves first compiling detections into a list of confidence scores and one-hot encoded ground truth labels. We then sweep through all available confidence scores to identify the optimal threshold. Here, we calculate the proportion of correct predictions by applying each threshold (see Figure \ref{fig:calibration}). The optimal threshold is identified by maximizing accuracy, which is the ratio of the true positive and true negative predictions. Additionally, to comprehensively evaluate model behavior, we compute Receiver Operating Characteristics (ROC) as shown in Figure \ref{fig:calibration}, by computing the true positive rate (TPR), and false positive rate (FPR) across all thresholds. Once we obtain the optimal threshold, we utilize it to calibrate the predictions from the VLM. We show the accuracy vs confidence plots before and after calibration in \Cref{fig:calibration}.

\section{Video Automaton Generation Function}
\label{sec:appx_va_construction}
Given a calibrated score set (see \Cref{eq:calibrated_score}) across all frames $\mathcal{F}_n$ (where $n$ is the frame index of the video) and propositions in $\propset$, we construct the video automaton $\va$ using the video automaton generation function (see \Cref{eq:va}).  
\begin{equation} 
    \label{eq:calibrated_score}
    C^\star = \{ C^\star_{p_i, j} \mid p_i \in \propset, j \in \{1, 2, \dots, n\} \}.
\end{equation}

As shown in \Cref{alg:av_generation}, we first initialize the components of the automaton, including the state set $Q$, the label set $\lambda$, and the transition probability set $\delta$, all with the initial state $q_0$. Next, we iterate over $C^\star$, incrementally constructing the video automaton by adding states and transitions for each frame. This process incorporates the proposition set and associated probabilities of all atomic propositions. We compute possible labels for each frame as binary combinations of $\propset$ and calculate their probabilities using the  $C^\star$.

\section{\neusv{} Prompt Suite}

    \paragraph{Creating the Dataset:}
        Our dataset is carefully designed to evaluate temporal fidelity and event sequencing in generated videos. It spans four themes -- ``Nature", ``Human \& Animal Activities", ``Object Interactions", and ``Driving Data", with each theme containing 40 prompts. These prompts vary in complexity, categorized into 20 basic, 15 intermediate, and five advanced prompts based on the number of temporal and logical operators (see \Cref{tab:benchmark-stats}). These prompts were generated using GPT-4o with the system prompt in \Cref{prompt:our-dataset}.
        To ensure quality, each prompt was manually verified for clarity, completeness, and relevance.

    \paragraph{Generating Videos from Prompts:}
        We generated videos for all 160 prompts using both open-source and closed-source models. For open-source models, we utilized pre-trained weights and code by cloning their HuggingFace Spaces and querying them using $\texttt{huggingface\_client}$. For closed-source models (Gen3 and Pika), we implemented a custom CLI that reverse-engineers their front-end interfaces to automate video generation requests. In total, we produced 640 videos (160 prompts $\times$ four models). We also plan to launch a publicly available leaderboard on HuggingFace after acceptance, which will allow continuous evaluation of new T2V models as they emerge.

\begin{figure}[t]
    \centering
    \includegraphics[width=0.589\linewidth]{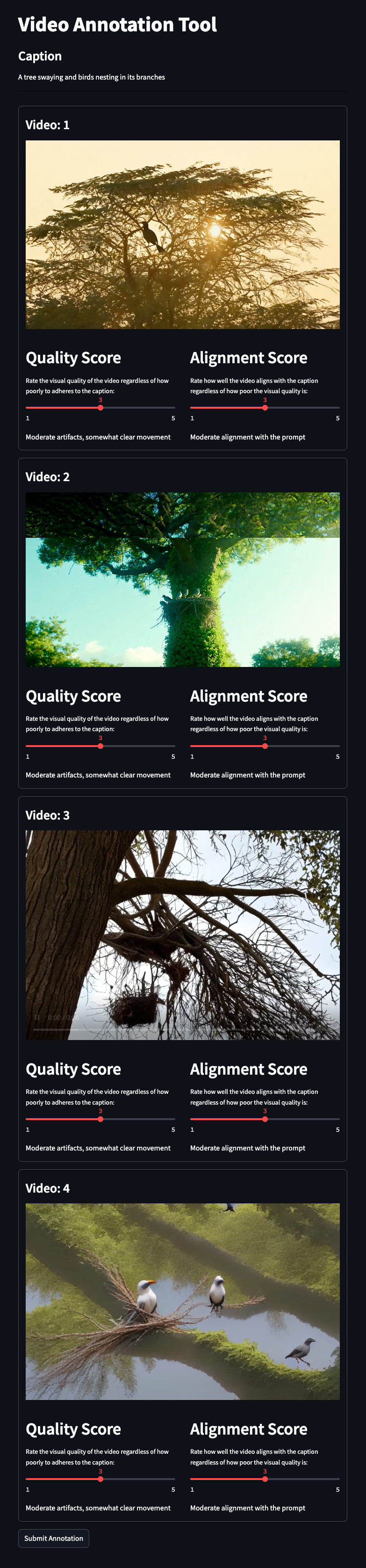}
    \caption{\textbf{Tool for Annotating Videos.} Subjects are instructed to disambiguate quality and alignment during annotation, scoring each from a range of 1 through 5.}
    \label{fig:annotation-tool}
\end{figure}

    \paragraph{Annotations:}
        Annotations were crowdsourced from 20 participants, including contributors from social media platforms like X (formerly Twitter) and LinkedIn. As shown in \Cref{fig:annotation-tool}, annotators were instructed to evaluate the alignment of videos with their corresponding prompts while disentangling visual quality from text-to-video alignment. 

    \paragraph{Insights into Prompts:}
        We showcase representative examples from our dataset in \Cref{tab:prompts-pt1} and \Cref{tab:prompts-pt2}, highlighting the diversity and complexity of prompts. These examples provide prompts to represent them in different modes. In the future, we plan to expand the dataset by adding more prompts across existing themes and introducing new categories to further enhance its utility and scope.

\begin{table}[!h]
    \centering
    \resizebox{\linewidth}{!}{
    \begin{tabular}{ccccc}
        \toprule
        \multirow{2}{*}{Theme} & \multicolumn{3}{c}{Complexity} & \multirow{2}{*}{Total Prompts} \\
        \cmidrule{2-4}
        & Basic & Intermediate & Advanced & \\
        
        \midrule
        Nature & 20 & 15 & 5 & 40 \\
        Human \& Animal Activities & 20 & 15 & 5 & 40 \\
        Object Interactions & 20 & 15 & 5 & 40 \\
        Driving Data & 20 & 15 & 5 & 40 \\
        
        \midrule
        \textbf{Total} & \textbf{80} & \textbf{60} & \textbf{20} & \textbf{160} \\
        \bottomrule
    \end{tabular}
    }
    \caption{\textbf{Statistics of \neusv{} Prompt Suite.} We include prompts from various themes across different complexities to evaluate T2V models on a total of 160 prompts.}
    \label{tab:benchmark-stats}
\end{table}

    \begin{algorithm*}[t]
    \DontPrintSemicolon
    \KwInput{Set of semantic score across all frames given all atomic propositions \{$C^\star = C^\star_{p_i, j} \mid p_i \in \propset, j \in \{1, 2, \dots, n\}$\}, set of atomic propositions $\propset$}
    \KwOutput{Video automaton $\va$}
    \Begin{
        $Q \leftarrow \{q_0\}$ \tcp*{Initialize the set of states with the initial state}
        
        $\lambda \leftarrow \{(q_0, \text{initial})\}$ \tcp*{Initialize the set of labels with the initial label}
        
        $\delta \leftarrow \{\}$ \tcp*{Initialize the set of state transitions}
        
        $Q_p \leftarrow \{q_0\}$ \tcp*{Track the set of previously visited states}
        
        $n \leftarrow \frac{|C^\star|}{|\propset|}$ \tcp*{Calculate the total number of frames $n$}
        
        \For{$j \gets 1$ \KwTo $n$}{
            $Q_c \leftarrow \{\}$ \tcp*{Track the set of current states}
            
            \For{$e_j^k \in 2^{|\propset|}$}{
                \tcp{\( e_j^k \): unique combination of 0s and 1s for atomic propositions in \( \propset \)}
                $\lambda(q_j^k) = \{v_1, v_2, \dots, v_i \mid v_i \in \{1,0\}, \forall i \in \{1, 2, \dots, |\propset|\}\}$ 

                $pr(j, k) \leftarrow 1$ \tcp*{Initialize probability for the label}
                
                \For{$v_i \in \lambda(q_j^k)$}{
                    \tcp{Calculate probability for $e_j^k$}
                    \If{$v_i = 1$}{
                        $pr(j, k) \leftarrow pr(j, k) \cdot C^\star_{p_i, j}$
                    }
                    \Else{
                        $pr(j, k) \leftarrow pr(j, k) \cdot (1 - C^\star_{p_i, j})$
                    }
                
                \tcp{Add state and define transitions if the probability is positive}
                \If{$pr(j, k) > 0$}{
                    $Q \leftarrow Q \cup \{q_{j}^k\}$
    
                    $Q_c \leftarrow Q_c \cup \{q_{j}^k\}$
                    
                    $\lambda \leftarrow \lambda \cup \{(q_{j}^k, \lambda(q_j^k))\}$
        
                    \For{$q_{j-1} \in Q_p$}{
                    $\delta(q_{j-1}, q_j^k) \leftarrow pr(j, k)$
                    
                    $\delta \leftarrow \delta \cup \{\delta(q_{j-1}, q_j^k)\}$}
                    
                    \EndFor
                    }
                    
                \EndFor
                }
            $Q_p \leftarrow Q_c$ \tcp*{Update previous state}
            
            \EndFor    
            }
        
        \EndFor
        }    
        \tcp{Add terminal state}
        $Q \leftarrow Q \cup \{q_{j+1}^0\}$
        
        $\lambda \leftarrow \lambda \cup \{(q_{j+1}^0, \text{terminal})\}$
    
        \For{$q_{j-1} \in Q_p$}{
        $\delta(q_{j-1}, q_{j+1}^0) \leftarrow 1$
        
        $\delta \leftarrow \delta \cup \{\delta(q_{j-1}, q_{j+1}^0)\}$}
        \EndFor
    \tcp{Return video automaton}
    $\va \leftarrow (Q, q_0, \delta, \lambda)$ \\
    \Return $\va$\;
    }
    \caption{Video Automaton Generation}
    \label{alg:av_generation}
    \end{algorithm*}
\clearpage
\begin{minipage}{\linewidth}
    \begin{tcolorbox}[
        title=Generating Temporally Extended Prompts (.md),
        colback=white,
        colframe=gray,
        colbacktitle=gray
        ]
        \ttfamily\scriptsize
        **Objective**: Generate individual prompts for a text-to-video generation benchmark. Each prompt should focus on specific temporal operators and adhere to the given theme and complexity level. Your goal is to create clear, vivid prompts that illustrate events occurring in time, with a strong emphasis on temporal relationships. \newline \smallskip

--- \newline \smallskip

\#\#\#\# **Instructions for Prompt Generation**

1. **Theme**: One of the following themes: Nature, Human and Animal Activities, Object Interactions, or Driving Data

2. **Complexity Level**: \\
   \hangindent=2em 
   - **Basic (1 Operator)**: Use only **one temporal operator** ("Always," "And," or "Until") in the prompt. \\
   - **Intermediate (2 Operators)**: Use **two temporal operators** in a sequence. The prompt should clearly connect the events with each operator in a natural, coherent way. \\
   - **Advanced (3 Operators)**: Use **three temporal operators** in a chain, showing a progression of events. Each part should flow logically to the next.

3. **Available Temporal Operators**: \\
   \hangindent=2em 
   - **"Always"**: Describes an event that continuously occurs in the background or context. \\
   - **"And"**: Combines two events happening simultaneously or in coordination. \\
   - **"Until"**: Describes an event that occurs until another event starts. \newline \smallskip

--- \newline \smallskip

\#\#\#\# **Examples**

\#\#\#\#\# **Theme: Nature**

- **Basic (1 Operator)**: \\ 
  \hangindent=2em 
  - "Always a river flowing gently." \\
  - "Rain pouring until the sun comes out." \\
  - "A bird chirping and a dog barking nearby."

\bigskip
\hspace{7em}$\cdots$ \\
\smallskip
\textbf{<~redacted more examples >} \\
\smallskip
\hspace{7em}$\cdots$

--- \newline \smallskip

**Ensure** that each prompt uses only the specified number of operators based on complexity, with clear temporal transitions between events. Use vivid language to create a realistic and engaging scenario. Try not to use language that is too abstract or ambiguous, focusing on concrete actions and events. Do not include any acoustic information in the prompts, as they are meant to describe visual scenes only. \newline \smallskip

I shall provide you with a theme, complexity level, and the number of prompts you need to generate. You must only output the prompts, each on a new line, without any additional information. \newline \smallskip

Are you ready? 
    \end{tcolorbox}
    \captionsetup{type=prompt}
    \captionof{prompt}{\textbf{System Prompt for \neusv{} Prompt Suite.} Used to query GPT-4o to generate temporally extended prompts across different themes and complexities.}
    \label{prompt:our-dataset}
\end{minipage}

\begin{table*}[t]
    \footnotesize
    \centering
    \begin{tabular}{C{0.6in}c|C{5.1in}}
        \toprule
        Theme & Complexity & Prompts and Specification Modes \\
        \midrule
        
        Nature & Basic & 
        \textbf{Prompt:} Snow falling until it covers the ground \newline
        \textbf{Object Existence:} (``snow") \until~(``ground") \newline
        \textbf{Object Action Alignment:} (``snow\_falls" \until~``ground\_is\_covered") \newline
        \textbf{Spatial Relationship:} F (``snow\_covers\_ground") \newline
        \textbf{Overall Consistency:} (``snow\_falling" \until~``it\_covers\_the\_ground") \\
        
        \cmidrule{3-3}
        & & 
        \textbf{Prompt:} Always waves crashing against the rocky shore \newline
        \textbf{Object Existence:} G (``waves" \& ``shore") \newline
        \textbf{Object Action Alignment:} G (``waves\_crash\_against\_rocky\_shore") \newline
        \textbf{Spatial Relationship:} G (``waves\_on\_shore") \newline
        \textbf{Overall Consistency:} G (``waves\_crashing\_against\_the\_rocky\_shore") \\
        
        \cmidrule{2-3}
        & Intermediate & 
        \textbf{Prompt:} The sun shining until the clouds gather, and then rain begins to fall \newline
        \textbf{Object Existence:} (``sun\_shining" \until~``clouds\_gather") \& F (``rain\_begins\_to\_fall") \newline
        \textbf{Object Action Alignment:} (``sun\_shines" \until~``clouds\_gather") \& F (``rain\_falls") \newline
        \textbf{Spatial Relationship:} G (``sun\_over\_horizon" \& ``dew\_on\_grass") \newline
        \textbf{Overall Consistency:} (``sun\_shining" \until~``clouds\_gather") \& F (``rain\_begins\_to\_fall") \\
        
        \cmidrule{3-3}
        & & 
        \textbf{Prompt:} A butterfly resting on a flower until a gust of wind comes, and then it flies away \newline
        \textbf{Object Existence:} (``butterfly" \& ``flower") \until~(``wind") \newline
        \textbf{Object Action Alignment:} (``butterfly\_rests\_on\_flower" \until~``gust\_of\_wind\_comes") \& F (``butterfly\_flies\_away") \newline
        \textbf{Spatial Relationship:} (``butterfly\_on\_flower") \until~(``butterfly\_flies\_away") \newline
        \textbf{Overall Consistency:} (``butterfly\_resting\_on\_a\_flower" \until~``gust\_of\_wind\_comes") \& F (``it\_flies\_away") \\
        
        \cmidrule{2-3}
        & Advanced & 
        \textbf{Prompt:} Always a river flowing quietly through the valley, until the sky darkens with storm clouds, and rain begins to pour heavily \newline
        \textbf{Object Existence:} G (``river" \& ``storm\_clouds") \newline
        \textbf{Object Action Alignment:} ((``river\_flows\_quietly" \until~``sky\_darkens")) \& F(``rain\_pours") \newline
        \textbf{Spatial Relationship:} G(``river\_flowing\_through\_valley") \until~(``sky\_darkens\_with\_storm\_clouds") \& F ``rain\_begins\_to\_pour\_heavily" \newline
        \textbf{Overall Consistency:} G ((``river\_flowing\_quietly\_through\_the\_valley") \until~``sky\_darkens\_with\_storm\_clouds") \& F ``rain\_begins\_to\_pour\_heavily" \\

        \midrule

        Human and Animal Activities & Basic & 
        \textbf{Prompt:} A dog barking until someone throws a ball \newline
        \textbf{Object Existence:} (``dog") \until~(``ball") \newline
        \textbf{Object Action Alignment:} (``dog\_barks" \until~``someone\_throws\_ball") \newline
        \textbf{Spatial Relationship:} G (``clock\_on\_mantle" \& ``fireplace\_beneath\_mantle") \newline
        \textbf{Overall Consistency:} (``dog\_barking" \until~``someone\_throws\_a\_ball") \\
        
        \cmidrule{3-3}
        & & 
        \textbf{Prompt:} A bird singing until it flies away to another branch \newline
        \textbf{Object Existence:} (``bird") \until~(``branch") \newline
        \textbf{Object Action Alignment:} (``bird\_sings" \until~``bird\_flies\_away\_to\_another\_branch") \newline
        \textbf{Spatial Relationship:} (``bird\_on\_branch") \until~(``bird\_on\_different\_branch") \newline
        \textbf{Overall Consistency:} (``bird\_singing" \until~``it\_flies\_away\_to\_another\_branch") \\
        
        \cmidrule{2-3}
        & Intermediate & 
        \textbf{Prompt:} A child building a sandcastle until the tide rises, and then they watch it wash away \newline
        \textbf{Object Existence:} (``child" \& ``sandcastle") \until~``tide" \newline
        \textbf{Object Action Alignment:} ((``child\_builds\_sandcastle" \until~``tide\_rises")) \& F(``child\_watches\_it\_wash\_away") \newline
        \textbf{Spatial Relationship:} (``child\_building\_sandcastle" \until~``tide\_rises") \newline
        \textbf{Overall Consistency:} (``child\_building\_a\_sandcastle" \until~``tide\_rises") \& F ``child\_watches\_sandcastle\_wash\_away" \\
        
        \cmidrule{3-3}
        & & 
        \textbf{Prompt:} Always a cat lounging on the porch, and butterflies fluttering around \newline
        \textbf{Object Existence:} G (``cat" \& ``butterflies") \newline
        \textbf{Object Action Alignment:} G((``cat\_lounges" \& ``butterflies\_flutter")) \newline
        \textbf{Spatial Relationship:} G(``cat\_lounging\_on\_porch" \& ``butterflies\_fluttering\_around") \newline
        \textbf{Overall Consistency:} G (``cat\_lounging\_on\_the\_porch" \& ``butterflies\_fluttering\_around") \\
        
        \cmidrule{2-3}
        & Advanced & 
        \textbf{Prompt:} A dog digging in the backyard, until its owner arrives, and then they play fetch together \newline
        \textbf{Object Existence:} (``dog" \& ``backyard") \until~(``owner" \& ``ball") \newline
        \textbf{Object Action Alignment:} (``dog\_digs\_in\_backyard" \until~``owner\_arrives") \& F (``dog\_plays\_fetch\_with\_owner") \newline
        \textbf{Spatial Relationship:} (``dog\_in\_backyard") \until~(``owner\_arrives") \newline
        \textbf{Overall Consistency:} (``dog\_digging\_in\_the\_backyard" \until~``its\_owner\_arrives") \& F (``they\_play\_fetch\_together") \\

        \bottomrule
    \end{tabular}
    \caption{\textbf{\neusv{} Prompt Suite:} Illustrative prompts and their detailed specifications (across all four modes) for varying complexities within the ``Nature" and ``Human and Animal Activities" themes.}
    \label{tab:prompts-pt1}
\end{table*}

\begin{table*}[t]
    \footnotesize
    \centering
    \begin{tabular}{C{0.6in}c|C{5.1in}}
        \toprule
        Theme & Complexity & Prompts and Specification Modes \\
        \midrule
        
        Object Interactions & Basic & 
        \textbf{Prompt:} A lamp glowing until it is turned off \newline
        \textbf{Object Existence:} (``lamp") \until~(!``lamp") \newline
        \textbf{Object Action Alignment:} (``lamp\_glows" \until~``lamp\_is\_turned\_off") \newline
        \textbf{Spatial Relationship:} (``cars\_passing\_by\_person") \newline
        \textbf{Overall Consistency:} (``lamp\_glowing" \until~``it\_is\_turned\_off") \\
        
        \cmidrule{3-3}
        & & 
        \textbf{Prompt:} A car engine running and the dashboard lights flashing \newline
        \textbf{Object Existence:} (``car\_engine" \& ``dashboard\_lights") \newline
        \textbf{Object Action Alignment:} (``car\_engine\_runs"\&``dashboard\_flashes") \newline
        \textbf{Spatial Relationship:} (``car\_engine\_running" \& ``dashboard\_lights\_flashing") \newline
        \textbf{Overall Consistency:} (``car\_engine\_running" \& ``dashboard\_lights\_flashing") \\
        
        \cmidrule{2-3}
        & Intermediate & 
        \textbf{Prompt:} Always a record player spinning a vinyl, and light glowing softly from a nearby lamp \newline
        \textbf{Object Existence:} G (``record\_player" \& ``lamp") \newline
        \textbf{Object Action Alignment:} G((``record\_player\_spins"\&``lamp\_glows")) \newline
        \textbf{Spatial Relationship:} G(``vinyl\_on\_record\_player" \& ``light\_glowing\_from\_nearby\_lamp") \newline
        \textbf{Overall Consistency:} G (``record\_player\_spinning\_a\_vinyl" \& ``light\_glowing\_softly\_from\_a\_nearby\_lamp") \\
        
        \cmidrule{3-3}
        & & 
        \textbf{Prompt:} A drone hovering in the air until it reaches its next waypoint, and then it continues to fly \newline
        \textbf{Object Existence:} (``drone") \until~(``waypoint") \newline
        \textbf{Object Action Alignment:} (``drone\_hovers\_in\_air" \until~``drone\_reaches\_next\_waypoint") \& F (``drone\_continues\_to\_fly") \newline
        \textbf{Spatial Relationship:} (``drone\_in\_air") \until~(``drone\_reaches\_waypoint") \newline
        \textbf{Overall Consistency:} (``drone\_hovering\_in\_air" \until~``drone\_reaches\_next\_waypoint") \& F (``drone\_continues\_to\_fly") \\
        
        \cmidrule{2-3}
        & Advanced & 
        \textbf{Prompt:} A lightbulb flickering intermittently, until the switch is turned off, and then the room is cast into darkness \newline
        \textbf{Object Existence:} (``lightbulb") \until~``switch" \newline
        \textbf{Object Action Alignment:} ((``lightbulb\_flickers" \until~``switch\_turned\_off")) \& F(``room\_darkens") \newline
        \textbf{Spatial Relationship:} (``lightbulb\_flickering" \until~``switch\_is\_turned\_off") \newline
        \textbf{Overall Consistency:} (``lightbulb\_flickering\_intermittently" \until~``switch\_is\_turned\_off") \& F ``room\_is\_cast\_into\_darkness" \\

        \midrule

        Driving Data & Basic & 
        \textbf{Prompt:} The vehicle moving forward until it reaches a stop sign \newline
        \textbf{Object Existence:} ``vehicle" \until~``stop\_sign" \newline
        \textbf{Object Action Alignment:} (``vehicle\_moves" \until~``vehicle\_reaches\_stop\_sign") \newline
        \textbf{Spatial Relationship:} (``vehicle\_moving\_forward" \until~``vehicle\_at\_stop\_sign") \newline
        \textbf{Overall Consistency:} (``vehicle\_moving\_forward" \until~``vehicle\_reaches\_a\_stop\_sign") \\
        
        \cmidrule{3-3}
        & & 
        \textbf{Prompt:} A motorcycle revving and a bus pulling up beside it \newline
        \textbf{Object Existence:} (``motorcycle" \& ``bus") \newline
        \textbf{Object Action Alignment:} (``motorcycle\_revs"\&``bus\_pulls\_up") \newline
        \textbf{Spatial Relationship:} (``motorcycle\_revving" \& ``bus\_pulling\_up\_beside") \newline
        \textbf{Overall Consistency:} (``motorcycle\_revving" \& ``bus\_pulling\_up\_beside") \\
        
        \cmidrule{2-3}
        & Intermediate & 
        \textbf{Prompt:} A traffic light turning red until pedestrians finish crossing, and then it shifts to green \newline
        \textbf{Object Existence:} (``traffic\_light") \until~``pedestrians" \newline
        \textbf{Object Action Alignment:} ((``traffic\_light\_is\_red" \until~``pedestrians\_finish\_crossing")) \& F(``traffic\_light\_turns\_green") \newline
        \textbf{Spatial Relationship:} (``traffic\_light\_turning\_red" \until~``pedestrians\_finish\_crossing") \newline
        \textbf{Overall Consistency:} (``traffic\_light\_turning\_red" \until~``pedestrians\_finish\_crossing") \& F ``traffic\_light\_shifts\_to\_green" \\
        
        \cmidrule{3-3}
        & & 
        \textbf{Prompt:} Always an electric vehicle charging at the station, and its driver reading a book nearby \newline
        \textbf{Object Existence:} G (``vehicle" \& ``driver") \newline
        \textbf{Object Action Alignment:} G((``electric\_vehicle\_charges"\&``driver\_reads")) \newline
        \textbf{Spatial Relationship:} G(``electric\_vehicle\_charging\_at\_station" \& ``driver\_reading\_book\_nearby") \newline
        \textbf{Overall Consistency:} G (``electric\_vehicle\_charging\_at\_the\_station" \& ``driver\_reading\_a\_book\_nearby") \\
        
        \cmidrule{2-3}
        & Advanced & 
        \textbf{Prompt:} A car driving through the city streets, until it encounters a construction zone, and then it reroutes to an alternate path \newline
        \textbf{Object Existence:} (``car" \& ``streets") \until~(``construction\_zone" \& ``path") \newline
        \textbf{Object Action Alignment:} (``car\_drives\_through\_city\_streets" \until~``car\_encounters\_construction\_zone") \& F (``car\_reroutes\_to\_alternate\_path") \newline
        \textbf{Spatial Relationship:} (``car\_on\_city\_streets") \until~(``car\_at\_construction\_zone") \newline
        \textbf{Overall Consistency:} (``car\_driving\_through\_city\_streets" \until~``it\_encounters\_a\_construction\_zone") \& F (``it\_reroutes\_to\_an\_alternate\_path") \\

        \bottomrule
    \end{tabular}
    \caption{\textbf{\neusv{} Prompt Suite (continued):} Illustrative prompts and their detailed specifications (across all four modes) for varying complexities within the ``Object Interactions" and ``Driving Data" themes.}
    \label{tab:prompts-pt2}
\end{table*}



\end{document}